\documentclass[runningheads]{llncs}

\usepackage{eccv}

\usepackage{eccvabbrv}

\usepackage{graphicx}
\usepackage{booktabs}
\usepackage{animate}
\usepackage{subcaption}
\usepackage{makecell}
\usepackage{multirow}
\usepackage{color}
\usepackage{xcolor}
\usepackage{colortbl}
\usepackage{wrapfig}
\usepackage{csquotes}
\usepackage{amsmath}

\usepackage{listings}
\usepackage{xcolor}
\lstdefinestyle{jsonStyle}{
  belowcaptionskip=1\baselineskip,
  breaklines=true,
  frame=L,
  xleftmargin=\parindent,
  language=Java,
  showstringspaces=false,
  basicstyle=\footnotesize\ttfamily,
  keywordstyle=\bfseries\color{green!40!black},
  commentstyle=\itshape\color{purple!40!black},
  identifierstyle=\color{blue},
  stringstyle=\color{orange},
}

\definecolor{color3}{rgb}{0.95,0.95,0.95}
\usepackage[accsupp]{axessibility}  % Improves PDF readability for those with disabilities.

\usepackage[pagebackref,breaklinks,colorlinks,citecolor=eccvblue]{hyperref}
\usepackage{orcidlink}
\usepackage{bbding}

\begin{document}

% ---------------------------------------------------------------
% TODO REVIEW: Replace with your title
% \title{RegionMaker: Object-aware Open-domain Regional Image Animation via Short Prompt} 
\title{Follow-Your-Click: 
Open-domain Regional Image Animation via Short Prompts}

% TODO REVIEW: If the paper title is too long for the running head, you can set
% an abbreviated paper title here. If not, comment out.
\titlerunning{Follow-Your-Click}

% TODO FINAL: Replace with your author list. 
% Include the authors' OCRID for the camera-ready version, if at all possible.
\author{Yue Ma\inst{1}\thanks{Equal contribution.} \and
Yingqing He$^{1*}$ \and
Hongfa Wang$^{2,3*}$ \and
Andong Wang\inst{2} \and 
Chenyang Qi\inst{1} \and \\
Chengfei Cai\inst{2} \and 
Xiu Li\inst{3} \and 
Zhifeng Li\inst{2} \and  
Heung-Yeung Shum\inst{1,3} \and \\
Wei Liu\inst{2}$^{(}$\Envelope$^)$ \and 
Qifeng Chen\inst{1}$^{(}$\Envelope$^)$}

% TODO FINAL: Replace with an abbreviated list of authors.
\authorrunning{Y. Ma et al.}
% First names are abbreviated in the running head.
% If there are more than two authors, 'et al.' is used.

% TODO FINAL: Replace with your institution list.
\institute{\textsuperscript{\rm 1}HKUST, 
\textsuperscript{\rm 2}Tencent, Hunyuan, \textsuperscript{\rm 3}Tsinghua Univerisity\\
\url{https://follow-your-click.github.io/}
}

\maketitle

\begin{center}{
    \captionsetup{type=figure}
    \vspace{-1em}
    \begin{tabular}{c@{\hspace{0.1em}}c@{\hspace{0.1em}}c@{\hspace{0.1em}}c@{\hspace{0.1em}}c@{\hspace{0.1em}}c@{\hspace{0.1em}}c@{\hspace{0.1em}}c@{\hspace{0.1em}}c}
  User Click & Output & User Click  & Output & User Click  & Output \\
 \includegraphics[width=0.16\linewidth]{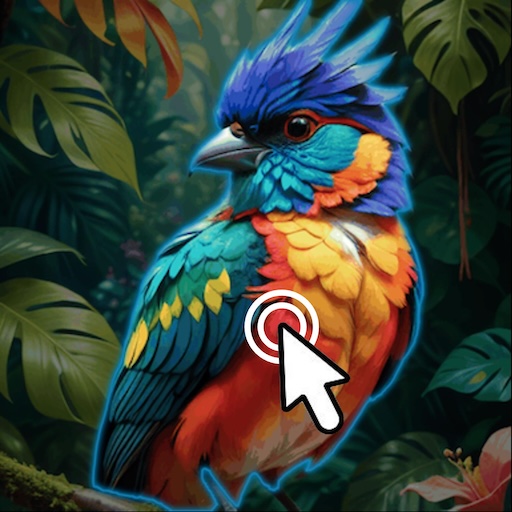}&
  \animategraphics[width=0.16\linewidth]{8}{gif/teaser/1/frame_}{1}{16} &   \includegraphics[width=0.16\linewidth]{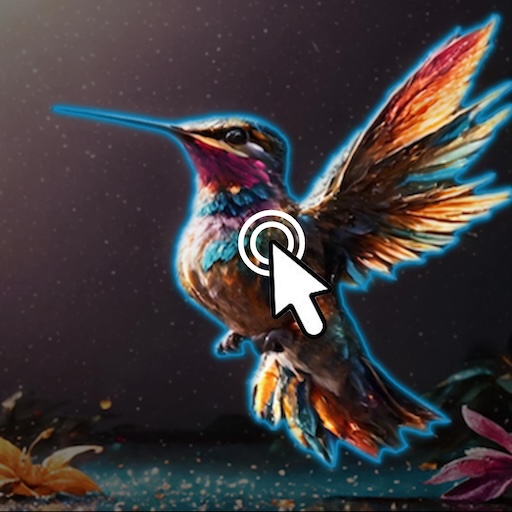}&
  \animategraphics[width=0.16\linewidth]{8}{gif/teaser/8/frame_}{1}{16} & \includegraphics[width=0.16\linewidth]{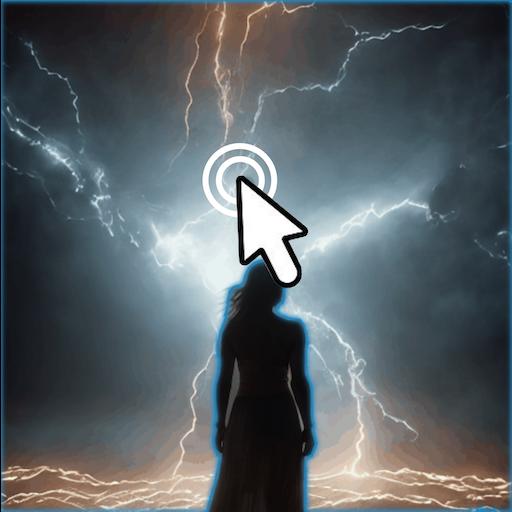}&
  \animategraphics[width=0.16\linewidth]{8}{gif/teaser/storm/frame_}{1}{16} \\
  \multicolumn{2}{c}{\textit{``\textcolor{red}{Tune} the head''}} & \multicolumn{2}{c}{\textit{``\textcolor{red}{Flap} the wings''}} & \multicolumn{2}{c}{\textit{``\textcolor{red}{Storm}''}}\\

    \includegraphics[width=0.16\linewidth]{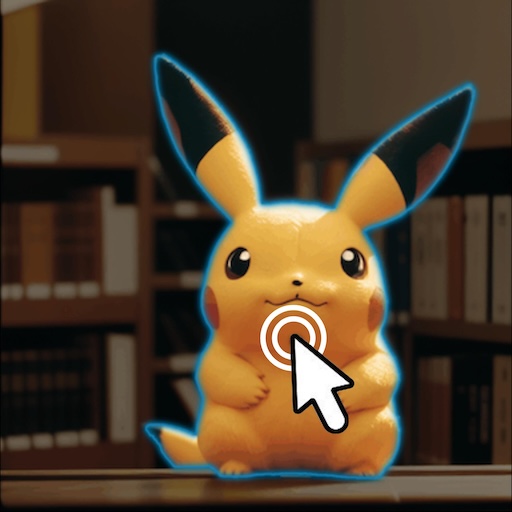}&
  \animategraphics[width=0.16\linewidth]{8}{gif/teaser/3/frame_}{1}{16} &   \includegraphics[width=0.16\linewidth]{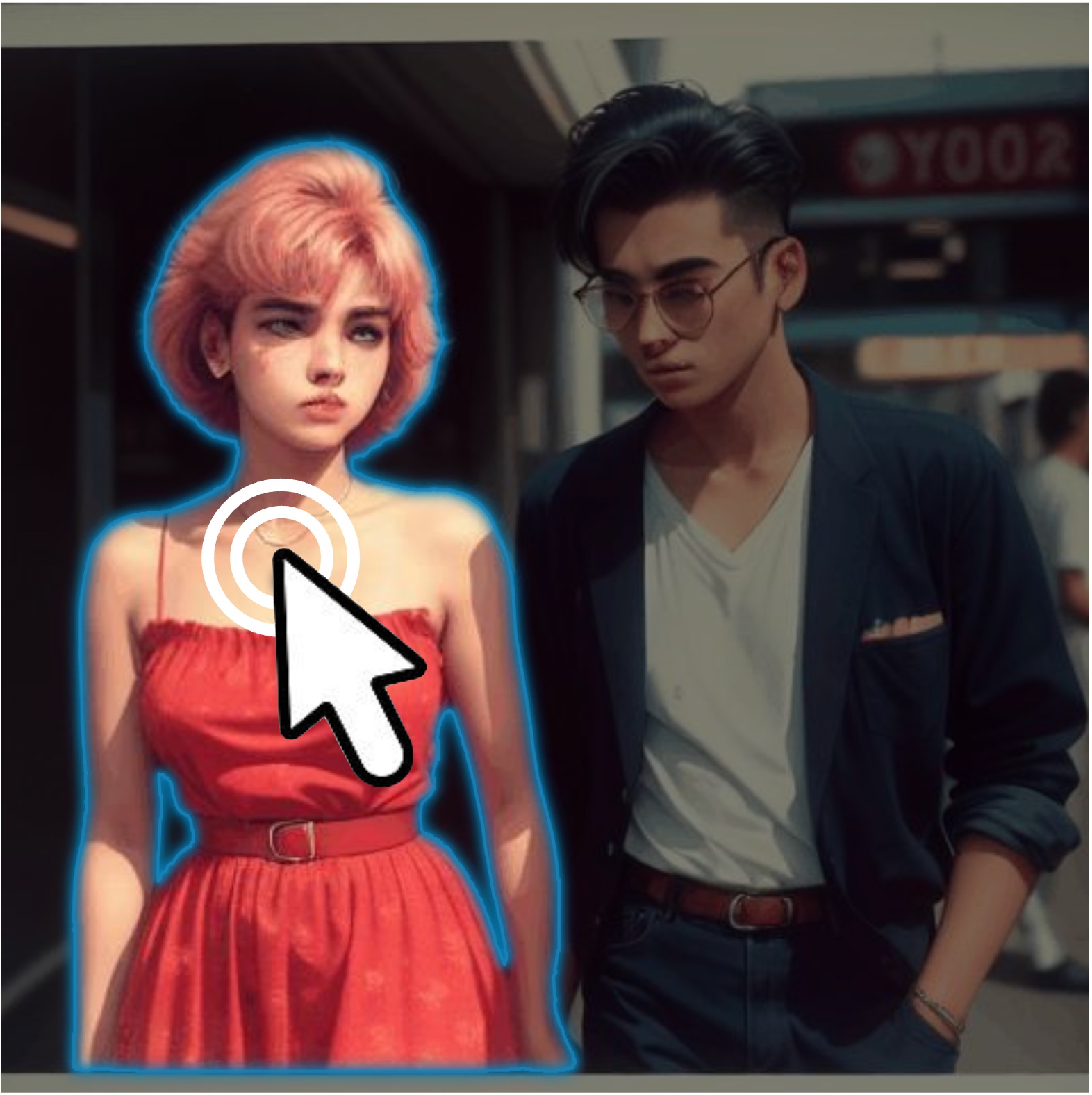}&
  \animategraphics[width=0.16\linewidth]{8}{gif/ablation_motion_cry/OFC_16/frame_}{1}{16} & \includegraphics[width=0.16\linewidth]{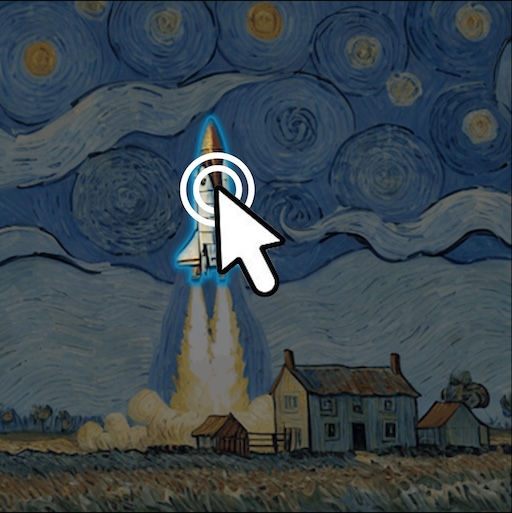}&
  \animategraphics[width=0.16\linewidth]{8}{gif/teaser/2/frame_}{1}{16}
  \\
  \multicolumn{2}{c}{\textit{``\textcolor{red}{Smile}''}} & \multicolumn{2}{c}{\textit{``\textcolor{red}{Sad}''}} & \multicolumn{2}{c}{\textit{``\textcolor{red}{Launch}''}}\\

  %   \includegraphics[width=0.16\linewidth]{images/teaser_7.jpg}&
  % \animategraphics[width=0.16\linewidth]{8}{gif/teaser/7/frame_}{1}{16} &   \includegraphics[width=0.16\linewidth]{images/teaser_2.jpg}&
  % \animategraphics[width=0.16\linewidth]{8}{gif/teaser/2/frame_}{1}{16} \\
  % \multicolumn{2}{c}{\textit{``\textcolor{red}{Drift}''}} & \multicolumn{2}{c}{\textit{``\textcolor{red}{Launch}''}}\\
  
\includegraphics[width=0.16\linewidth]{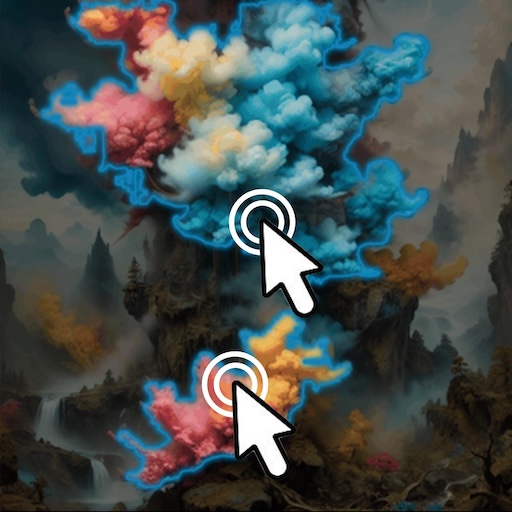}&
  \animategraphics[width=0.16\linewidth]{8}{gif/teaser/7/frame_}{1}{16} &
   \includegraphics[width=0.16\linewidth]{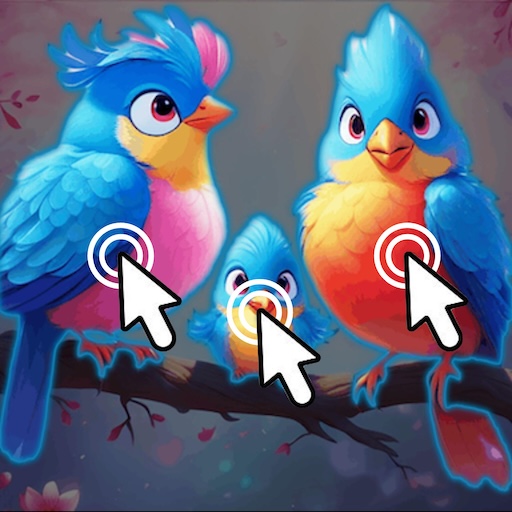}&
  \animategraphics[width=0.16\linewidth]{8}{gif/teaser/5/frame_}{1}{16} &   \includegraphics[width=0.16\linewidth]{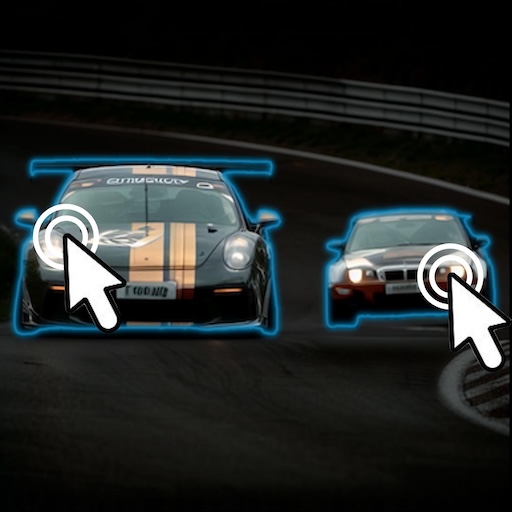}&
  \animategraphics[width=0.16\linewidth]{8}{gif/teaser/6/frame_}{1}{16} \\
  \multicolumn{2}{c}{\textit{``\textcolor{red}{Drift}''}} & \multicolumn{2}{c}{\textit{``\textcolor{red}{Dancing}''}} & \multicolumn{2}{c}{\textit{``Drive \textcolor{red}{back} and \textcolor{red}{forward}''}}\\
    \end{tabular}
    \captionof{figure}{
    \textbf{Regional Image Animation using a Click and a Short Prompts}. 
    We present a novel framework that facilitates locally aware image animation via a user-provided click (\textit{where} to move) and a short motion prompt (\textit{how} to move).
    Our framework can provide vivid object movement, background movement (e.g., storm), and multiple object movements.
    %
    % We present RegionMaker, a regional open-domain image animation framework controlled by user clicks and short motion prompts.
    % an efficient framework for open-domain image regional animation using motion-related prompts. 
    \textit{Best viewed with Acrobat Reader}, which supports clicking on the video to play the animations.
    % \textit{for all the videos}. 
    % \textbf{
    \textit{Static frames and videos of all results are provided in supplementary materials}.
    % }
    \label{fig:teaser}
}
}\end{center}

\begin{abstract}
Despite recent advances in image-to-video generation, better controllability and local animation are less explored.
Most existing image-to-video methods are not locally aware and tend to move the entire scene.
However, human artists may need to control the movement of different objects or regions.
Additionally, current I2V methods require users not only to describe the target motion but also to provide redundant detailed descriptions of frame contents.
These two issues hinder the practical utilization of current I2V tools.
In this paper, we propose a practical framework, named Follow-Your-Click, to achieve image animation with a simple user click (for specifying \textit{what} to move) and a short motion prompt (for specifying \textit{how} to move).
Technically, we propose the first-frame masking strategy, which significantly improves the video generation quality, and a motion-augmented module equipped with a short motion prompt dataset to improve the short prompt following abilities of our model.
To further control the motion speed, we propose flow-based motion magnitude control to control the speed of target movement more precisely.
Our framework has simpler yet precise user control and better generation performance than previous methods.
% Besides the simple controlling mechanism, our method also has high generation performance and can deliver high-quality and diverse image animation results.
%
Extensive experiments compared with 7 baselines, including both commercial tools and research methods on 8 metrics, suggest the superiority of our approach.

\end{abstract}

\section{Introduction}
\label{sec:intro}
Image-to-video generation (I2V) aims to animate an image into a dynamic video clip with reasonable movements. 
It has widespread applications in the filmmaking industry, augmented reality, and automatic advertising. 
Traditionally, image animation methods mainly focus on domain-specific categories, such as natural scenes~\cite{cheng2020time, jhou2015animating, li2023generative, xiong2018learning}, human hair~\cite{xiao2023automatic}, portraits~\cite{geng2018warp, wang2022latent} and bodies~\cite{bertiche2023blowing, blattmann2021understanding, weng2019photo, karras2023dreampose}, limiting their practical application in real world. 
In recent years, the significant advancements in the diffusion models~\cite{rombach2022high, saharia2022photorealistic, nichol2021glide} trained on large-scale image datasets have enabled the generation of diverse and realistic images based on text prompts.
Encouraged by this success, researchers have begun extending these models to the realm of I2V, aiming to leverage the strong image generation priors for image-to-video generation~\cite{xing2023dynamicrafter, wang2024videocomposer, shi2024motion, chai2023stablevideo}.

However, existing I2V works~\cite{chai2023stablevideo, xing2023dynamicrafter, wang2023videocomposer, i2vgenxl} have a lack of control over which part of the image needs to be moved, and they produce videos with the movement of the entire scene; And some works such as SVD~\cite{chai2023stablevideo} tend to deliver videos always with camera movement, ignoring the more vivid object movement. They cannot achieve regional image animation which is important to human artists (\eg, the user may want to animate the foreground object while keeping the background static). 
Besides, the typical prompts that users provide to I2V models are the descriptions of the entire scene contents.
However, the spatial content is fully described via the input image which is not necessary for users to describe it again. 
In fact, a more intuitive way is to provide motion-only prompts, but current approaches are less sensitive to short motion prompts.
% due to they are trained on long prompt captions, and the captions are processed as a whole without a particular focus on motion-related verbs.
%
A common hypothesis in previous works is that the diffusion model is a prompt-driven framework, and a detailed prompt may enhance the quality of the generated results. 
However, such a feature dramatically limits the practical application for users in the real world.  
The existing datasets such as WebVid~\cite{bain2021frozen} and HDVILA~\cite{xue2022advancing} mainly focus on describing scenes and events in their captions, while ignoring the motion of the objects. 
Training on such datasets may result in a decrease in the quality of generated motion and insensitivity towards motion-related keywords.

In this paper, we aim to devise a more practical and controllable I2V model that can address such problems.
To this end, we propose \textbf{Follow-Your-Click}, a novel I2V framework that is capable of regional image animation via a user click and following short motion prompts.
To achieve this simple user interaction mechanism while obtaining good generation performance, we first simply integrate SAM~\cite{cheng2023segment} to convert user clicks to binary regional masks, which serve as one of our network conditions.
Then to better learn the temporal correlation correctly, we introduce an effective \textit{first-frame masking} strategy and observe a large margin of performance gains.
To achieve the short prompt following abilities, we construct a dataset referred to as \textit{WebVid-Motion}, which is built by leveraging a large language model (LLM) for filtering and annotating the video captions, emphasizing human emotion, action, and common motion of objects.
We then design a \textit{motion-augmented module} to better adapt to the dataset and enhance the model’s response to motion-related words and understand short prompt instructions.
%
% To accurately identify movable objects and their corresponding movable regions within an image, we introduce motion area guidance using optical flow during the training stage, allowing for control of the motion area mask. 
Furthermore, we also observe that different object types may exhibit varied motion speeds. In previous works~\cite{xing2023dynamicrafter}, frame rate per second (FPS) primarily serves as a global scaling factor to indirectly adjust the motion speed of multiple objects. However, it cannot effectively control the speed of moving objects. For instance, a video featuring a sculpture may have a high FPS but zero motion speed. 
To enable accurate learning of motion speed, we propose a novel \textit{flow-based motion magnitude control}.
% we propose a novel motion strength strategy using the magnitude of optical flow

With our design, we achieve remarkable results on eight various evaluation metrics.
Our method can also facilitate the control of \textit{multiple} object and moving types via multiple clicks.
Besides, it is easy to integrate our approach with controlling signals, such as human skeletons, to achieve a more fine-grained motion control.
%
% In summary, our contributions are the following three-fold:
Our contributions can be summarized as follows:
\begin{itemize}
 \renewcommand{\labelitemi}{\textbullet}
\item To the best of our knowledge, Follow-Your-Click is the first framework supporting a simple \textit{click} and \textit{short motion prompt} for regional image animation. 

\item To achieve such a user-friendly and controllable I2V framework, technically, we propose the \textit{first-frame masking} to enhance the general generation quality, a \textit{motion-augmented module} with an equipped \textit{short prompt dataset} for short prompt following, and a \textit{flow-based motion magnitude} for a more accurate motion speed control.

\item We conducted extensive experiments and user studies to evaluate our approach, which shows our method achieves state-of-the-art performance.
% , in terms of generation quality, text-video alignment, mask-video alignment, and temporal consistency.
\end{itemize}
% This design empowers users to manipulate object motions with very sparse trajectory annotations

% contributions:

\section{Related Work}
\subsection{Text-to-Video Generation}
Text-to-video generation is a popular topic with extensive research in recent years. 
Before the advent of diffusion models,
many approaches have developed based on transformer architectures~\cite{ramesh2021zero, yu2022scaling, yu2021vector, wang2024animatelcm, ding2022cogview2, yan2021videogpt, hong2022cogvideo, ma2023magicstick, ho2022imagen, he2023weaklysupervised, xiao2023bridging,he2023camouflaged} to achieve textual control for generated content. The emergency of diffusion models~\cite{song2020denoising} delivers
higher quality and more diverse results. Early works such as LVDM~\cite{he2022latent} and modelscope~\cite{wang2023modelscope} explore the integration of temporal modules. Video diffusion model (VDM)~\cite{ho2022video} is proposed to model low-resolution videos using a spacetime factorized U-Net in pixel space.
% DALLE-2~\cite{ramesh2022hierarchical} enhances text-image alignments by leveraging the representation space of CLIP~\cite{radford2021learning} and Imagen~\cite{saharia2022photorealistic} employs cascaded diffusion models to achieve high-definition image generation. 
Recent models beneﬁt from the stability of training diffusion-based model~\cite{rombach2022high}. 
These models can be scaled by a huge dataset and show surprisingly good results on text-to-video generation. Magic-video~\cite{zhou2022magicvideo} and gen1~\cite{gen2} initialize the model from text-to-image~\cite{rombach2022high} and generate the continuous contents through extra time-aware layers. Additionally, a category of VDMs that decouples the spatial and temporal modules has emerged~\cite{guo2023animatediff, guo2023sparsectrl}. While they provide the potential to control appearance and motion separately, they still face the challenge of video regional control. 

Even though these models can produce high-quality videos, they mainly rely on textual prompts for semantic guidance, which can be ambiguous and may not precisely describe users' intentions. To address such a problem, many control signals such as structure~\cite{esser2023structure, xing2023make, gao2023imperceptible}, pose~\cite{ma2023follow, zhang2023controlvideo, wang2023gen}, and Canny edge~\cite{zhang2023controlvideo} are applied for controllable video generation. Many recent and concurrent methods in Dynamicrafter~\cite{xing2023dynamicrafter}, VideoComposer~\cite{wang2023videocomposer}, and I2VGen-XL~\cite{i2vgenxl} explore RGB images as a condition to guide video synthesis. However, they concentrate on a certain domain and fail to generate temporally coherent frames and realistic motions while preserving details of the input image. Besides, most of the prompts are used to describe the image content, users can not animate the image according to their intent. Our approach is based on text-conditioned VDMs and leverages their powerful generation ability to animate the objects in the images while preserving the consistency of background.

\subsection{Image Animation}
Image-to-video generation involves an important demand: maintaining the identity of the input image while creating a coherent video. This presents a significant challenge in striking a balance between preserving the image's identity and the dynamic nature of video generation.
Early approaches based on physical simulation~\cite{dorkenwald2021stochastic, wang2022latent, wang2023gen, prashnani2017phase, siarohin2021motion, he2024strategic} concentrate on simulating the movement of certain objects, result in poor generalizability because of the separate modeling of each object category. 
With the success of deep learning, more GAN-based works~\cite{hinz2021improved,karras2020analyzing, shaham2019singan} get rid of manual segmentation and can synthesize more natural motion. Mask-based approaches such as MCVD~\cite{voleti2022mcvd} and SEINE~\cite{chen2023seine} predict future video frames starting from single images to achieve the task. They play a crucial role in preserving the consistency of the input image's identity throughout the generated video frames, ensuring a smooth transition from static to dynamic. Currently, mainstream works based on diffusion~\cite{holynski2021animating, mallya2022implicit, weng2019photo, chen2023controlavideo, gao2024inducing} can generate frames using the video diffusion model. Dynamicrafter~\cite{xing2023dynamicrafter} and Livephoto~\cite{chen2023livephoto} propose a powerful framework for real image animation and achieve a competitive performance. The plug-to-play adapters such as I2V-adapter~\cite{guo2023i2v} and PIA~\cite{zhang2023pia} apply public Lora~\cite{civitai} weights and checkpoints to animate an image. But they only focus on the curated domain and fail to generate temporally coherent real frames. 
Additionally, Some commercial large-scale models,  Gen-2~\cite{gen2}, Genmo~\cite{genmo}, and Pika Labs~\cite{pikalabs} deliver impressive results in the realistic image domain in its November 2023 update. 
However, these works cannot achieve regional image animation and accurate control. 
Among the concurrent works, the latest version of Gen-2 released the motion brush in January 2024, which supports regional animation. However, It still faces the challenge of synthesizing realistic motion (see Fig.~\ref{fig:Qualitative comparison}).
Additionally, it cannot support the user click and short prompt interactions.
Furthermore, as a commercial tool, Gen-2 will not release technical solutions and checkpoints for research.  
In contrast, our method holds unique advantages in its simple interactions, motion-augmented learning, and better generation quality.

\section{Preliminaries}
\label{sec:preliminary}
\noindent\textbf{Latent Diffusion Models (LDMs).}
% \paragraph{Latent Diffusion Models (LDMs)} 
We choose Latent Diffusion Model~\cite{rombach2022high}  (LDM) as the backbone generative model.
Derived from Diffusion Models, LDM reformulates the diffusion and denoising procedures within a latent space. 
This process can be regarded as a Markov chain, which incrementally adds Gaussian noise to the latent code.
First, an encoder $\mathcal{E}$ compresses a pixel space image $x$ to a low-resolution latent $z=\mathcal{E}(x)$
% , such that the latent 
, which can be reconstructed from latent feature to image $ \mathcal{D}(z) \approx x $ by decoder $\mathcal{D}$.
Then, a U-Net~\cite{ronneberger2015u} $\varepsilon_\theta$ 
% composed of 
with self-attention~\cite{vaswani2017attention} and cross-attention is trained to estimate the added noise via this objective:
% \vspace{-0.5em}
\begin{equation}
\label{eq:mse}
\min _\theta E_{z_0, \varepsilon \sim N(0, I), t \sim \text { Uniform }(1, T)}\left\|\varepsilon-\varepsilon_\theta\left(z_t, t, p \right)\right\|_2^2,
\end{equation}
% \vspace{-0.2em}
where $p$ is the embedding of the text prompt and $z_t$ is a noisy sample of $z_0$ at timestep $t$. After training, we can generate a clean image latent $z_0$ from random Gaussian noises $z_T$ and text embedding $p$ through step-by-step denoising and then decode the latent into pixel space by $\mathcal{D}$.

\noindent\textbf{Video latent diffusion models (VDMs).}
Following the previous works~\cite{rombach2022high, guo2023animatediff}, we expand the latent diffusion model to a video version (VDM) by incorporating the temporal motion module. In detail, the weights of spatial modules in VDMs are initialized with the pre-trained image LDMs and are frozen during training. This operation could help the model to inherit the generative priors from the powerful image LDM. The temporal motion modules, which comprise 1-D temporal attention, are inserted after each spatial module and they are responsible for capturing the temporal dependencies between representations of the same spatial location across different frames. Given a video $\mathbf{x} \in \mathbb{R}^{L \times C \times H \times W }$ where $L, C, H, W$ represent the video length, number of channels, height and width respectively, we first encode it into a latent space frame-by-frame, obtaining a video latent $\mathbf{z}$ where $\mathbf{z} \in \mathbb{R}^{L \times c \times h \times w }$.
Then, both the forward diffusion process and backward denoising process are performed in this latent space. Finally, the generated videos are obtained through the decoder.

% \section{Method}

\section{Follow-Your-Click}
\subsection{Problem Formulation}
Given a still image, our goal is to animate user-selected regions, creating a short video clip that showcases realistic motion while keeping the rest of the image static. 
% objects contents
Formally, given an input image $\mathcal{I}$, 
% a region mask $\mathcal{M}$, 
a point prompt $p$, 
and a short motion-related verb description of the desired motion $t$, our approach produces a target animated video $\mathcal{V}$.
% Formally, given an input image $\mathcal{I}$, a point prompt $p$, and a short verb description of the desired motion $t$, RegionMaker produces a target animated video $\mathcal{V}$.
We decompose this task into several sub-problems including improving the generation quality of local-aware regional animation, achieving short motion prompt controlled generation, and motion magnitude controllable generation. 
Note that the target region is utilized for selecting the animated object rather than limiting the motion of the generated object in subsequent frames. In other words, the object is not constrained to remain within the specified areas and can move outside of them if necessary.

\begin{figure}[tb]
  \centering
  \includegraphics[width=\linewidth]{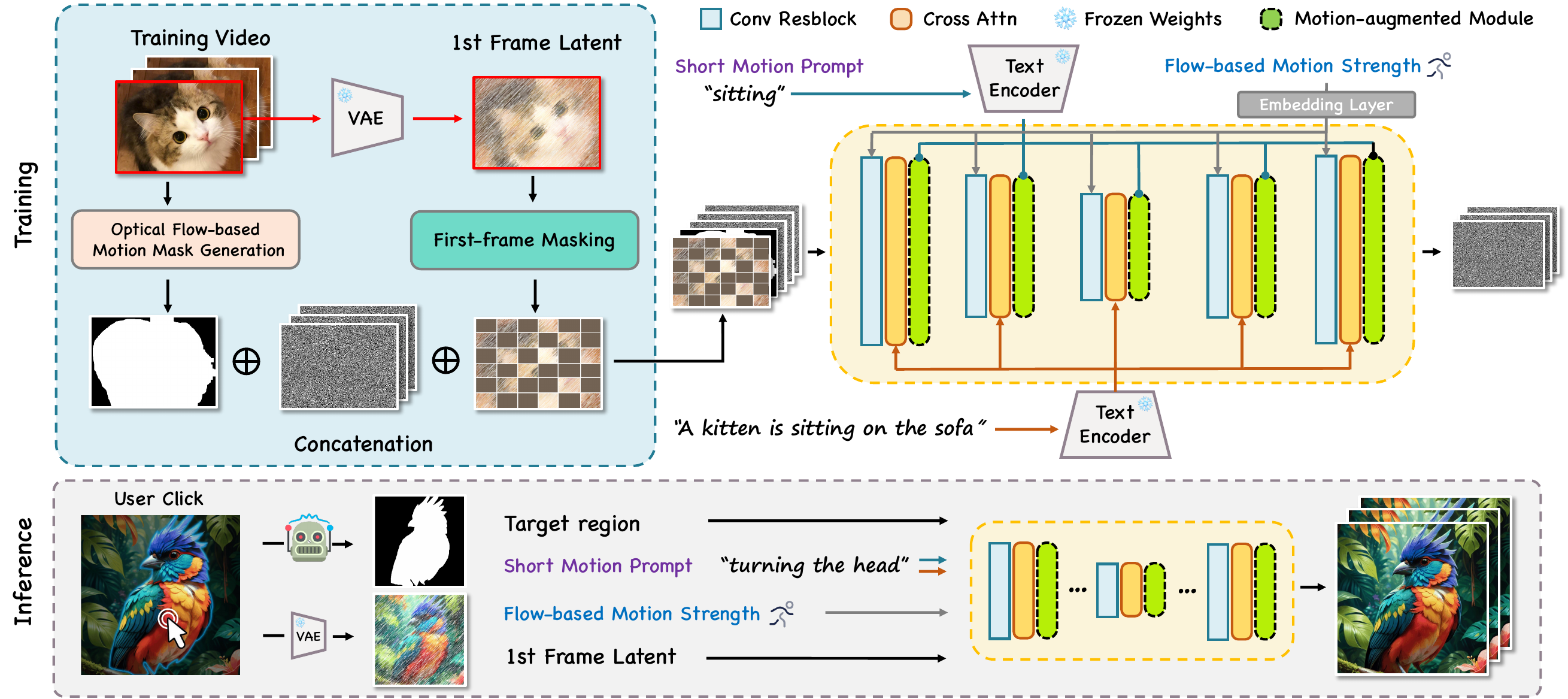}
  % \vspace{-0.6cm}
  \caption{
  \textbf{Framework overview.} 
  The key components of our framework are the first-frame masking, motion-augmented module for short motion prompt following, and flow-based motion strength control.
  During inference, the regional animation can be achieved by user clicks and short motion prompts.
  }

  \label{fig:Framework}
\end{figure}

\vspace{-0.6cm}
% \noindent\textbf{Optical flow for motion area guidance.} 
\subsubsection{User Interaction and Control.}

Given an input image that the user wants to animate.
An intuitive way is first to choose which part of the image needs to move, then use the text prompt to describe the desired moving pattern.
Current approaches, such as research works I2VGen-XL, SVD, dynamicrater, and commercial tools like Pika Lab and  Genmo, lack the ability of regional control. 
The motion brush of Gen-2~\cite{gen2} and animate-anything~\cite{dai2023animateanything} can achieve such a goal but the motion mask needs to be provided or drawn by users, which is not efficient and intuitive for users.
Thus, to provide a user-friendly control, we design to use a \textit{point prompt} instead of a binary mask.
Furthermore, current image-to-video methods require the input prompt to describe the entire scene and frame content, which is tedious and unnecessary. 
On the contrary, we simplify this procedure with a short motion prompt, using only the verb word or short phrase.
To achieve this, we integrate a promptable segmentation tool SAM~\cite{cheng2023segment} to convert the point to prompt $p$ to a high-quality object mask $\mathcal{M}$. The masked-controlled regional animation will be introduced in~\cref{sec:regional_animation}.
To achieve the short prompt following, we propose a motion-augmented module described in~\cref{sec:motion_aug_module}.

\subsection{Regional Image Animation}
\label{sec:regional_animation}

\noindent\textbf{Optical flow-based motion mask generation.} 
Training on public datasets such as WebVid~\cite{bain2021frozen} and HDVILA~\cite{xue2022advancing} directly is challenging to achieve regional image animation due to the lack of corresponding binary mask guidance for regions with large movement. 
To solve this issue, we utilize the optical flow prediction model to automatically generate the mask indicating the moving regions.
Specifically, give training video frames $\{x_0, x_1 ..., x_{L-1}\}$, we utilize an open-sourced optical flow estimator $\mathcal{E}_{{flow}}$~\cite{teed2020raft} to extract the optical flow map $\mathcal{F}_{i}$ of each two consecutive frame pairs, where $i$ is the frame index of the video.
For each flow map $\mathcal{F}_{i}$, we threshold the map into a binary one $\mathcal{M}_i$ via a threshold calculated via its average magnitude.
Finally, we take the union of all masks {$\mathcal{M}_1, \mathcal{M}_2, ..., \mathcal{M}_{L-1}$} to get the final mask $\mathcal{M}_{final}$ to represent area of motion. 
Formally, the motion area guidance is implemented as

\begin{equation}
\label{equ:auronorm}
\begin{gathered}
\mathcal{F}_{i} = \mathcal{E}_{{flow}}( {x}_{i}, {x}_{i-1}), \\
\mathcal{M}_{i} = \text{Binarize}(\mathcal{F}_{i}, \text{Avg}(\left \| \mathcal{F}_{i} \right \| )), \ 
\mathcal{M}_{final} = \bigcup_{i=0}^{L-1}(\mathcal{M}_{i}).
\end{gathered}
\end{equation}
where $i = 1, 2, 3, \hdots , L$, $\text{Binarize}(\cdot, \cdot )$ is the binarization operation and 
% $\mathcal{F}_{i}$ is the optical flow of $i$-th frames, and 
$\left \| \cdot  \right \| $ denotes magnitude of optical flow in each pixel. 
During training, we use $\mathcal{M}_{final}$ to represent the motion area of ground truth videos.
% As shown in Fig.~\ref{fig:Framework}, downsampled binary mask $\mathcal{M}_{final}$ will be concatenated with the masked input image latent and clean Gaussian noise in channel dimension, and then fed into the network for training. 
During inference, we transfer the user clicks into the binary mask via the promptable image segmentation tool SAM~\cite{cheng2023segment} and then feed the binary mask to our network.
We also study the generalization ability of conditional masks in supplementary materials.

\noindent\textbf{First-frame masking training.}
After obtaining the moving region mask $\mathcal{M}_{final}$, we concatenate the downsampled version, the first frame latent ${z}_{0}$, and random noise in the channel dimension in the latent space, obtaining input with size $[9, L, h, w]$ and then fed it into the network. 
${z}_{0}$ is the latent of the first frame $x_0$ which is encoded via the VAE encoder $\mathcal{E}$.
The $\mathcal{M}_{final}$ is downsampled to match the resolution of the frame latent.
The mask of the target generated frame {$\mathcal{M}_1, \mathcal{M}_2, ..., \mathcal{M}_{L-1}$} is set to zero, and the first frame serves as guidance and is repeated to $L$ frames.
The $9$ channels consist of $4$ channels of input image latent, $4$ channels of the generated frames, and $1$ channel of the binary mask.
We adopt the $\mathbf{v}$-prediction parameterization proposed in \cite{salimans2022progressive} for training since it has better sampling stability when a few of the inference steps.
%
% Then we insert a new motion-augmented module for optimization parameters (will introduced in \ref{sec:motion_aug_module}).
However, we observe that training directly in this manner exhibits temporal structure distortion issues.
% \noindent\textbf{Masking input image latent.} 
Inspired by the recent masked strategy works~\cite{he2022masked, feichtenhofer2022masked, ma2022simvtp}, we hypothesize that augmenting the condition information in training can help the model to learn the temporal correlation better.
%
% Specifically, during the training stage, we select the first frame ${x}_{0} $ of the video clip as input image $\mathcal{I}$, which then is projected to latent space using VAE encoder $\mathcal{E}$. 
Therefore, we randomly mask the latent embedding of the input image $z_0$ by a ratio of $\mathcal{R}$, setting the masked region to 0.
As shown in Fig.~\ref{fig:Framework}, the masked first frame latent, along with the downsampled $\mathcal{M}_{final}$ and noisy video latent $\mathbf{z}$, are concatenated and fed into the network for optimization.
% Following the original objective in AnimateDiff~\cite{guo2023animatediff}, we calculate the MSE loss between the ground truth video clip and the generated video clip.
Empirically, we discover that randomly masking the input image latent can significantly improve the quality of the generated video clip.
In Sec.~\ref{sec:ab}, we conduct a detailed analysis of the selection of mask ratio. 
% Note that we choose to apply such a mask strategy in the latent space is that it helps reduce computational cost. 

\subsection{Temporal Motion Control}
% \subsection{Temporal Motion Control via short prompt}
% \subsection{Motion-augmented module for short prompt following}
\label{sec:motion_aug_module}
% \noindent\textbf{Motion-augmented module.}
% In this section, we introduce our devised motion-augmented module.

\noindent\textbf{Short motion caption construction.}
We discover that captions in current extensive datasets always comprise numerous scene descriptive terms alongside fewer dynamic or motion-related descriptions.
To enable the achieve better short prompt following, we construct the WebVid-Motion dataset, a dataset by filtering and re-annotating the WebVid-10M dataset using GPT4~\cite{gpt4}.
In particular, we construct 50 samples to achieve in-context learning of GPT4. 
Each sample contains the original prompt, objects, and their short motion-related descriptions. 
These samples are fed into GPT4 in JSON format, and then we ask the same question to GPT4 to predict other short motion prompts in WebVid-10M. 
Finally,  the re-constructed dataset contains captions and their motion-related phrases, such as \enquote{tune the head}, \enquote{smile}, \enquote{blink} and \enquote{running}. 
We finetune our model on this dataset to obtain a better ability of short motion prompt following.
% get a better ability of short motion prompt following.

% \noindent\textbf{Short motion prompt injection.}
\noindent\textbf{Motion-augmented module.}
With a trained model via the previous techniques~\cite{guo2023animatediff}, to make the network further aware of short motion prompts, 
we design the motion-augmented module to improve the
model’s responses to motion-related prompts. 
In detail, we insert a new cross-attention layer in each motion module block.
The short motion-related phrases are fed into a motion-augmented module for training, and during inference, these phrases are input into both the motion-augmented module and the cross-attention module in U-Net.
% During the other parameters in the model are kept frozen.
% We enhance video latent diffusion model’s performance on text alignment with the proposed motion-augmented cross attention and keep its other modules as is. 
% Firstly, 
% During training stage, the cross attention layer between the representation and short motion-related word embedding is injected into the first motion block in each module to enhance the motion.  
%
Thanks to this module, our model can generate the desired performance during inference with just a short motion-related prompt provided by the user, eliminating the need for redundant complete sentences.

\noindent\textbf{Optical flow-based motion strength control.}
The conventional method for controlling motion strength primarily relies on adjusting frames per second (FPS) and employs the dynamic FPS mechanism during training \cite{zhou2022magicvideo}. 
However, we observe that the relationship between motion strength and FPS is not linear. 
Due to variations in video shooting styles, there can be a significant disparity between FPS and motion strength. For instance, even in low-FPS videos (where changes occur more \textit{rapidly} than in high-FPS videos), slow-motion videos may exhibit minimal motion. 
This approach fails to represent the intensity of motion accurately.
To address this, we propose using the magnitude of optical flow as a means of controlling the motion strength. 
As mentioned in~\cref{sec:regional_animation}, once we obtain the mask for the area with the most significant motion, we calculate the average magnitude of optical flow within that region. 
This magnitude is then projected into positional embedding and added to each frame in the residual block, ensuring a consistent application of motion strength across all frames.

% \subsection{Discussions on Motion Control using Short prompt}

% 这里是否需要写在webvid-motion 上finetune
\section{Experiments}
In this section, we introduce our detailed implementation in Sec.~\ref{sec:implementation}. 
Then we evaluate our approach with various baselines to comprehensively evaluate our performance in Sec.~\ref{sec:compare}. 
% Then we evaluate our method with both commercial tools and research works to comprehensively evalute our performance in \ref{sec:compare}. 
We then ablate our key components to show their effectiveness in Sec.~\ref{sec:ab}.
Finally, we provide two applications to demonstrate the potential of integrating our approach with other tools in Sec.~\ref{sec:application}.

\subsection{Implementation Details}
\label{sec:implementation}
In our experiments, the spatial modules are based on Stable Diffusion (SD) V1.5~\cite{rombach2022high}, and motion modules use the corresponding AnimateDiff~\cite{guo2023animatediff} checkpoint V2. 
We freeze the SD image autoencoder to encode each video frame to latent representation individually. We train our model for 60k steps on the WebVid-10M~\cite{bain2021frozen} and then finetune it for 30k steps on the reconstructed WebVid-Motion dataset. 
The training videos have a resolution of $512 \times 512$ with 16 frames and a stride of 4.
% The sixteen consecutive frames at the resolution of $512 \times 512$ from the input video are sampled for temporal consistency learning and the stride is set to 4. 
The overall framework is optimized with Adam~\cite{loshchilov2017decoupled} on 8 NVIDIA A800 GPUs for three days with a batch size of 32. 
We set the learning rate as $1 \times 10^{-4}$ for better performance. 
The mask ratio of the first frame is 0.7 during the training process. At inference, we apply DDIM sampler~\cite{song2020denoising} with classifier-free guidance~\cite{ho2022classifier} scale 7.5 in our experiments.
% for regional image animation.

\subsection{Comparison with baselines}
\label{sec:compare}
\begin{figure*}[!t]
\vspace{-0.3cm}
  % \centering
  \begin{tabular}{c@{\hspace{0.1em}}c@{\hspace{0.1em}}c@{\hspace{0.1em}}c@{\hspace{0.1em}}c@{\hspace{0.1em}}c@{\hspace{0.1em}}c}
  User Click & Gen-2~\cite{gen2} & Genmo~\cite{genmo} & Pika Labs~\cite{pikalabs} & {\small Animate-A}~\cite{dai2023animateanything} \\
  % & {\small I2VGen-XL}~\cite{i2vgenxl} & {\small RegionMaker(ours)}\\
    \includegraphics[width=0.2\linewidth]{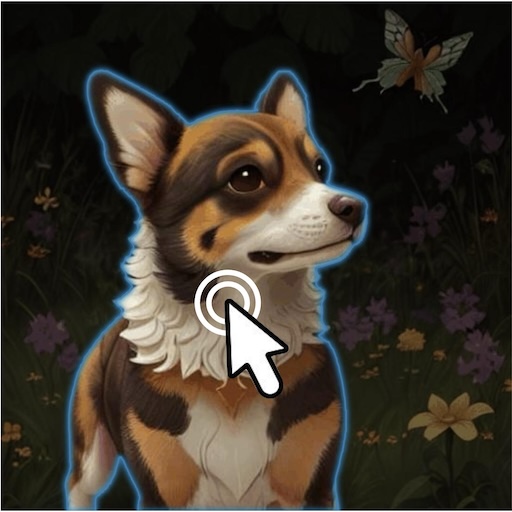} &
    \animategraphics[width=0.2\linewidth]{8}{gif/comparison_dog/dog_gen/frame_}{1}{16}&
     \animategraphics[width=0.2\linewidth]{8}{gif/comparison_dog/dog_genmo/frame_}{1}{16}&
     \animategraphics[width=0.2\linewidth]{8}{gif/comparison_dog/dog_pika/frame_}{1}{12}&
     \animategraphics[width=0.2\linewidth]{8}{gif/comparison_dog/dog_aa/frame_}{1}{16}
     \\
      \textit{``Shake body''}  & SVD~\cite{blattmann2023stable} & Dynamic~\cite{xing2023dynamicrafter} & I2VGen-XL~\cite{pikalabs} & {\small Ours} \\
  % & {\small I2VGen-XL}~\cite{i2vgenxl} & {\small RegionMaker(ours)}\\
    &
    \animategraphics[width=0.2\linewidth]{8}{gif/comparison_dog/dog_svd/frame_}{1}{10}&
     \animategraphics[width=0.2\linewidth]{8}{gif/comparison_dog/dog_dynamicrafter/frame_}{1}{16}&
     \animategraphics[width=0.2\linewidth]{8}{gif/comparison_dog/dog_i2v/frame_}{1}{16}&
     \animategraphics[width=0.2\linewidth]{8}{gif/comparison_dog/ours/frame_}{1}{16}
    \end{tabular}
\vspace{-0.7em}
  \caption{\textbf{Qualitative comparisons between baselines and our approach}.
  We compare with both close-sourced commercial tools including Gen-2~\cite{gen2}, Genmo~\cite{genmo}, and Pika~\cite{pikalabs} and research works including Animate-anything~\cite{dai2023animateanything}, SVD\cite{chai2023stablevideo}, Dynamicrafter\cite{xing2023dynamicrafter}, and I2VGen-XL \cite{i2vgenxl}.
  Please click the video to play the animated clips via \textit{Adobe Acrobat Reader}. 
  \textit{Static frames are provided in supplementary materials}.
  }
  \vspace{-0.5cm}
\label{fig:Qualitative comparison} 
\end{figure*}

\noindent\textbf{Qualitative results.}
We qualitatively compare our approach with the most recent open-sourced state-of-the-art animation methods, including Animate anything~\cite{dai2023animateanything}, SVD~\cite{blattmann2023stable}, Dynamicrafter~\cite{xing2023dynamicrafter} and I2VGen-XL~\cite{i2vgenxl}. 
We also compare our approach with commercial tools such as Gen-2~\cite{gen2}, Genmo~\cite{genmo}, and Pika Labs~\cite{pikalabs}. Note that the results we accessed on Feb.15th, 2024 might differ from the current product version due to rapid version iterations.
Dynamic results can be found in Fig.~\ref{fig:Qualitative comparison}. 
Given the benchmark images, their corresponding prompts, and selected regions, it can be observed that the videos generated by our approach exhibit better responses to short motion-related prompts \texttt{\enquote{Shake body}}. 
Meanwhile,
our approach achieves regional animation while also obtaining better preservation of details from the input image content.
In contrast, SVD and Dynamicrafter struggle to produce consistent video frames, as subsequent frames tend to deviate from the initial frame due to inadequate semantic understanding of the input image.
I2VGen-XL, on the other hand, generates videos with smooth motion but loses image details. 
We observe that Genmo is not sensitive to motion prompts and tends to generate videos with small motion. 
Animate-anything can achieve regional animation and generate motions as large as those produced by our approach, but it suffers from severe distortion and text alignment. 
As commercial products, 
Pika Labs and Gen-2 can produce appealing high-resolution and long-duration videos. However, Gen-2 suffers from the less responsive to the given prompts. Pika Labs tends to generate still videos with less dynamic and exhibits blurriness when attempting to produce larger dynamics.
These results verify that our approach has superior performance in generating consistent results using short motion-related prompts even in the presence of large motion.

\begin{table*}[!t]\footnotesize
% \vspace{-0.2cm}
\caption{\textbf{Quantative comparisons between baselines and our approach}. 
Our method demonstrates the best or comparable performance across multiple metrics.
% and achieves comparable results to the best methods in other metrics. 
The metrics for the best-performing method are highlighted in \textcolor{red}{red}, while those for the second-best method are highlighted in \textcolor{blue}{blue}.}
\resizebox{\columnwidth}{!}{
    % \center
    \centering
    \begin{tabular}{l | c c c c | c c c c}
    \toprule
    \rowcolor{color3}&  \multicolumn{4}{c}{Automatic Metrics } & \multicolumn{4}{c}{User Study}\\
    \rowcolor{color3} Method & $I_1$-MSE$\downarrow$ & Tem-Consis$\uparrow$ & Text-Align$\uparrow$ & FVD $\downarrow$ & Mask-Corr$\downarrow$ & Motion$\downarrow$ & Appearance$\downarrow$ & Overall $\downarrow$  \\
    % \hline
    % AnimateDiff~\cite{guo2023animatediff} & 105.9 &  0.6326 & 0.3293 &  0.8061 & 0.3916 & \\
    \hline \hline
    Gen-2~\cite{gen2} & $54.72$ & $0.8997$ & $0.6337$  & $496.17$& $3.12$ & $5.11$ & $2.52$ & $2.91$ \\
    Genmo~\cite{genmo} & $91.84$ & $0.8316$ & $0.6158$  & $547.16$ & $6.43$ & $4.57$ & $3.51$ & $3.76$ \\
    Pika Labs~\cite{pikalabs} & $\mathbf{\color{blue}33.27}$ & $\mathbf{\color{red}0.9724}$ & $\mathbf{\color{blue}0.7163}$  & $\mathbf{\color{blue}337.84}$ & $3.92$ & $\mathbf{\color{blue}2.86}$ & $\mathbf{\color{blue}2.17}$ & $\mathbf{\color{blue}2.88}$ \\
    \hline 
    Dynamicrafter~\cite{xing2023dynamicrafter} & $98.19$ & $0.8341$ & $0.6654$  & $486.37$ &  $5.27$ & $6.25$ & $4.91$ & $5.93$ \\
    I2VGen-XL~\cite{i2vgenxl} & $117.86$ & $0.6479$ & $0.5349$  & $592.13$ &  $7.19$ & $7.79$ & $6.98$ & $7.26$ \\
    SVD~\cite{i2vgenxl} & $43.57$ & $0.9175$ & $0.5007$  & $484.26$ & $4.91$ & $3.74$ & $3.94$ & $4.81$ \\
    Animate-anything~\cite{i2vgenxl} & $53.72$ & $0.7983$ & $0.6372$  & $477.42$ & $\mathbf{\color{blue}2.73}$ & $4.73$ & $5.47$ & $5.75$ \\
    \hline
    \textbf{Ours} & $\mathbf{\color{red}21.46}$ & $\mathbf{\color{blue}0.9613}$ & $\mathbf{\color{red}0.7981}$  & $\mathbf{\color{red}271.74}$ & $\mathbf{\color{red}1.38}$ & $\mathbf{\color{red}1.91}$ & $\mathbf{\color{red}1.87}$ & $\mathbf{\color{red}1.78}$ \\
    \bottomrule
    \end{tabular}
}
% \vspace{-0.7cm}
\label{table:compare_all}
\end{table*}

% }

\noindent\textbf{Quantitative results.} 
\label{sec:qr}
% There are a few image-to-video benchmarks, but they are limited to specific domains. 
For extensive evaluation, We construct a benchmark for quantitative comparison, which includes 30 prompts,
% \todo{Add number of samples of 4 automatic metrics} downloaded from the copyright-free website Pixabay. We.
images and corresponding region masks. The images are downloaded from the copyright-free website Pixabay and we use GPT4 to generate prompts for the image content and possible motion. The prompts and images encompass various contents (characters, animals, and landscapes) and styles (\textit{e.g.}, realistic, cartoon style, and Van Gogh style). 
Four evaluation metrics are applied to finish the quantitative test.
(1) \textbf{$I_{1}-$MSE}: We follow ~\cite{xing2023dynamicrafter} to measure the consistency between the generated first frame and the given image.
(2) \textbf{Temporal Consistency (Tem-Consis)}: It evaluates the temporal coherence of the generated videos. We calculate the cosine similarity between consecutive generated frames in the CLIP embedding space to measure the temporal consistency. 
(3) \textbf{Text alignment (Text-Align)}: We measure the degree of semantic alignment between the generated videos and the input short motion prompt.
Specifically, we calculate the similarity scores between the prompt and each generated frame using their features extracted by CLIP text and image encoders respectively.
(4) \textbf{FVD}: We report the Frechet Video Distance~\cite{unterthiner2019fvd} to evaluate the overall generation performance on 1024 samples from MSRVTT~\cite{xu2016msr}.
% (3) \textbf{Text-Sim}: We calculate the similarity between the prompt and each generated frame using their features extracted by CLIP text and image encoders respectively.
(5) \textbf{User Study}: We perform user study on four different aspects. 
\textit{Mask-Corr} assesses the correspondence of regional animation and guided mask.  \textit{Motion} evaluates the quality of generated motion. 
\textit{Appearance} measures the consistency of the generated 1st frame with a given image and \textit{Overall} evaluates the subjective quality of the generated videos. We ask 32 subjects to rank different methods in these four aspects. From Table.~\ref{table:compare_all}, It can be observed that our approach achieves the best video-text alignment and temporal consistency against baselines. As for the user study, our approach obtains the best performance in terms of temporal coherence and input conformity compared to commercial products, while exhibiting superior motion quality.

\begin{figure}[tb]
  \centering
  \includegraphics[width=\linewidth]{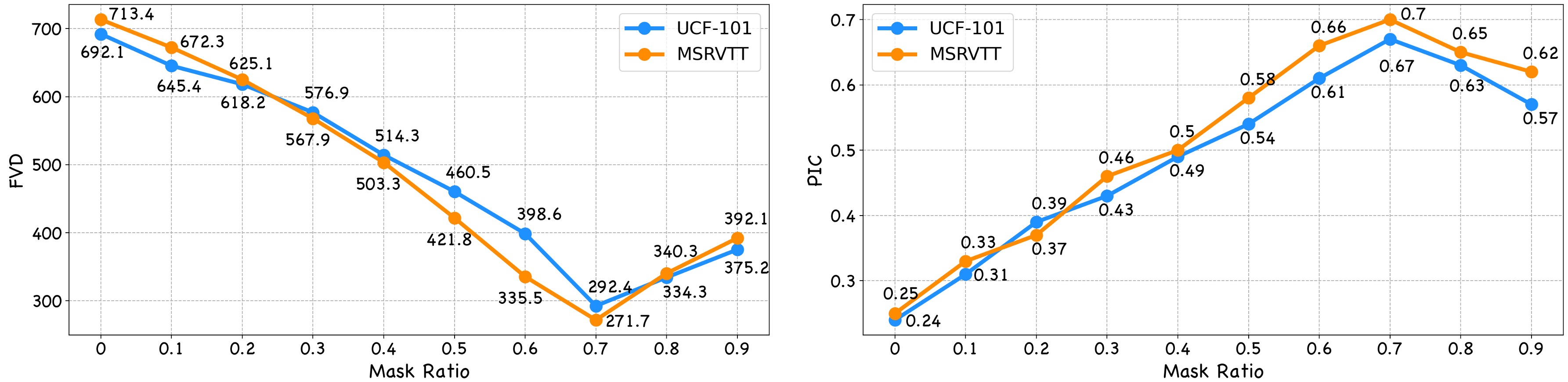}
  % \vspace{-0.6cm}
  \caption{
  \textbf{Ablation study about the masking ratio of the first-frame masking strategy.}
  Different masking ratios significantly affect the generation quality (FVD) and the perceptual input conformity (PIC)~\cite{xing2023dynamicrafter}.
  }
  \vspace{-0.4cm}
  \label{fig:ab masking ratio}
\end{figure}

\subsection{Ablation Study}
\label{sec:ab}
\begin{figure}[!t]
    % \centering
    \begin{minipage}[t]{0.6\textwidth}
        \centering
        \begin{tabular}{c@{\hspace{0.1em}}c@{\hspace{0.1em}}c@{\hspace{0.1em}}c@{\hspace{0.1em}}c@{\hspace{0.1em}}c@{\hspace{0.1em}}c@{\hspace{0.1em}}}
        User Click & Ratio=0 & Ratio=0.7  \\
      % & {\small I2VGen-XL}~\cite{i2vgenxl} & {\small RegionMaker(ours)}\\
        \includegraphics[width=0.3\linewidth]{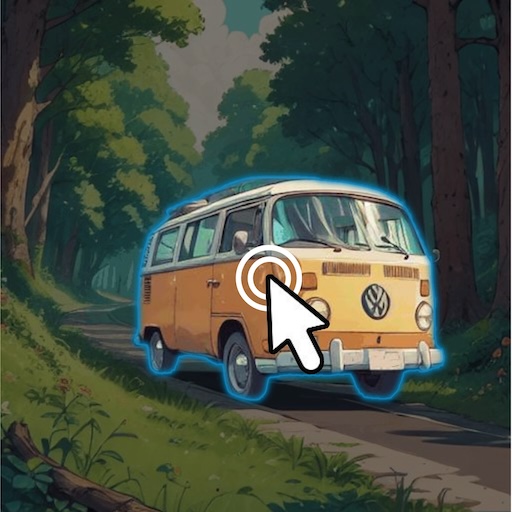} &
        \animategraphics[width=0.3\linewidth]{8}{gif/masked_vis/mask_0/frame_}{1}{16}& 
        \animategraphics[width=0.3\linewidth]{8}{gif/masked_vis/mask_70/frame_}{1}{16}
        % \\
        % \multicolumn{3}{c}{\textit{``driving''}}
        \end{tabular}
    \end{minipage}
    \begin{minipage}[t]{0.4\textwidth}
        \vspace{-33pt}
        \centering
        \caption{
        \textbf{Visual results of ablating different masking ratios.} Training without masking presents poor movement, temporal consistency and video quality. The prompt is \enquote{driving}.
        % Training with masking ratio 0.7 shows a large improved quality.
        % \textit{Best viewed with Acrobat Reader. Click the video to play the animation clips. \textbf{Static frames are provided in supplementary materials.}}
        }
    \end{minipage}
\vspace{-0.9cm}
\label{fig:limitation} 
\end{figure}

% \begin{wraptable}[8]{l}{12.5em}
%     \centering
%         \begin{tabular}{c@{\hspace{0.1em}}c@{\hspace{0.1em}}c@{\hspace{0.1em}}c@{\hspace{0.1em}}c@{\hspace{0.1em}}c@{\hspace{0.1em}}c@{\hspace{0.1em}}}
%         Input Image & Ratio=0 & Ratio=0.7  \\
%       % & {\small I2VGen-XL}~\cite{i2vgenxl} & {\small RegionMaker(ours)}\\
%         \includegraphics[width=0.2\linewidth]{images/car_click.jpg} &
%         % \includegraphics[width=0.14\linewidth]{gif/AZvsOthers/line3/gen2/011_0.jpg}&
%         \animategraphics[width=0.2\linewidth]{8}{gif/masked_vis/mask_0/frame_}{1}{16}& 
%         \animategraphics[width=0.2\linewidth]{8}{gif/masked_vis/mask_70/frame_}{1}{16}
%         \\
%         \multicolumn{3}{c}{\textit{``driving''}}
%         \end{tabular}
%         \caption{
%         \textbf{Visual results of ablating different masking ratios.} Training without masking present poor movement, temporal consistency and video quality.
%         Training with masking ratio 0.7 shows a large improved quality.
%         % \textit{Best viewed with Acrobat Reader. Click the video to play the animation clips. \textbf{Static frames are provided in supplementary materials.}}
%         }
% \label{fig:limitation} 
% \end{wraptable}

\begin{table}[t]
\caption{
\textbf{Quantitative ablation results of the motion-augmented module (MA) and our constructed short prompt dataset (Data)}. 
The best-performing methods are highlighted in \textcolor{red}{red}, and the second-best methods are highlighted in \textcolor{blue}{blue}.}
% The metrics for the best-performing method are highlighted in \textcolor{red}{red}, while those for the second-best method are highlighted in \textcolor{blue}{blue}.}
% \vspace{-1cm}
\centering
\resizebox{\linewidth}{!}{
\begin{tabular}{l|c c c c| c c c c}
\toprule
\rowcolor{color3}&  \multicolumn{4}{c}{Automatic Metrics } & \multicolumn{4}{c}{User Study}\\
 \rowcolor{color3} Method & $I_1$-MSE$\downarrow$ & Tem-Consis$\uparrow$ & Text-Align$\uparrow$ & FVD $\downarrow$ & Mask-Corr$\downarrow$ & Motion$\downarrow$ & Appearance$\downarrow$ & Overall $\downarrow$  \\
    \hline
    \hline
    w/o Data \& MA  & 35.72 & 0.8465 & 0.3659 & 698.21 & 2.92 & 3.27 & 3.34 & 3.18\\
    w/o MA  & $\mathbf{\color{blue}26.46}$ & $\mathbf{\color{blue}0.9178}$ & $\mathbf{\color{blue}0.6294}$ & $\mathbf{\color{blue}391.47}$ & $\mathbf{\color{blue}1.97}$ & $\mathbf{\color{blue}2.17}$ & $\mathbf{\color{blue}2.08}$ & $\mathbf{\color{blue}2.24}$ \\
    w/o Data  & 29.18 & 0.8824 & 0.4356 & 562.33 &  2.46 & 2.38 & 2.35 & 2.79 \\
    \hline
    Ours & $\mathbf{\color{red}21.46}$ & $\mathbf{\color{red}0.9613}$ &  $\mathbf{\color{red}0.7981}$ & $\mathbf{\color{red}271.74}$ & $\mathbf{\color{red}1.43}$ & $\mathbf{\color{red}1.59}$ & $\mathbf{\color{red}1.17}$ & $\mathbf{\color{red}1.31}$ \\
\bottomrule
\end{tabular}}

\label{tab:ablation motion augmented module}
% \end{wraptable}
\end{table}

\begin{figure}[tb]
\vspace{-0.5cm}
  \centering
  \includegraphics[width=\linewidth]{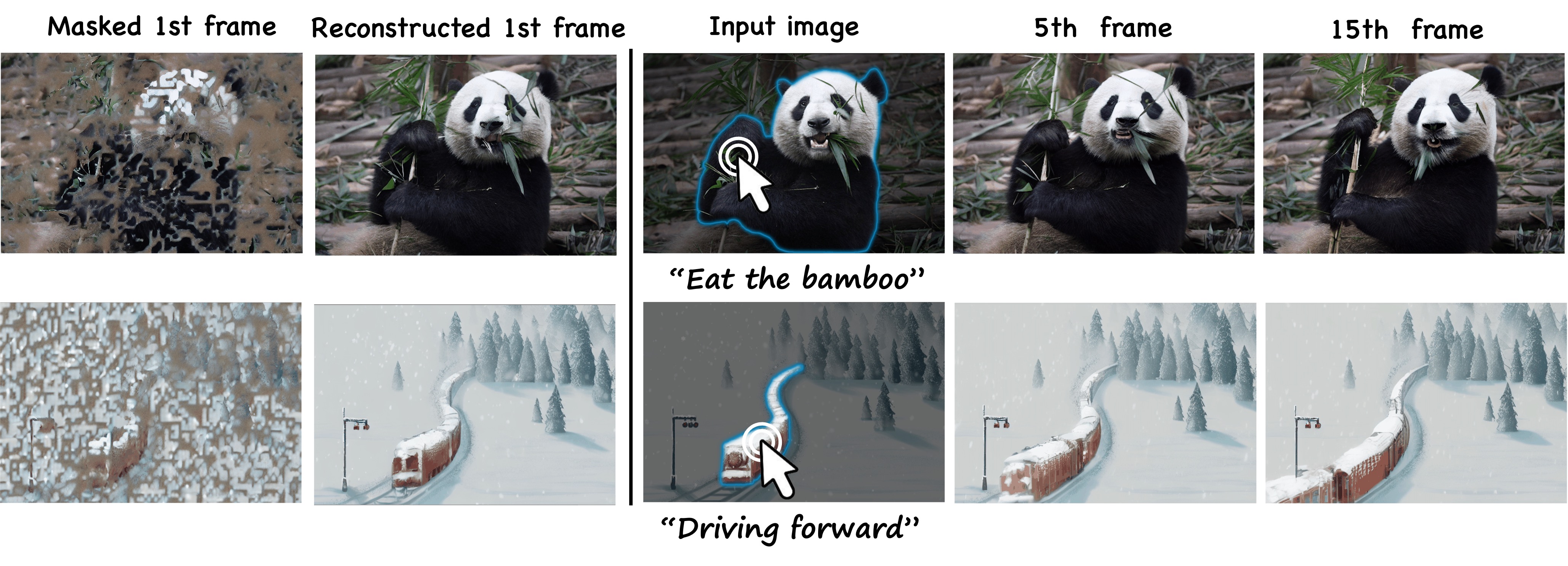}
  \vspace{-0.6cm}
  \caption{\textbf{Reconstruction and generation results of the masked first frame.}To clearly illustrate the performance of our reconstruction, we present static frames, while {\textit{dynamic videos are provided in the supplementary materials}}. }
  \vspace{-0.4cm}
  \label{fig:Reconstructions}
\end{figure}

% Although we have proved the effectiveness of the proposed strategies in Fig.~\ref{fig:Qualitative comparison} using two video examples, here, we ablate these designs in the video.

\noindent\textbf{Input image mask ratio.} To investigate the influence of the first frame masking strategy and different mask ratios for the input image in training, we conduct quantitative experiments varying the mask ratio from 0 to 0.9. 
Following ~\cite{xing2023dynamicrafter, blattmann2023align}, we evaluate the generation performance of all the methods on UCF-101~\cite{soomro2012ucf101} and MSRVTT~\cite{xu2016msr}. 
The Frechet Video Distance (FVD)~\cite{unterthiner2019fvd} and Perceptual Input Conformity (PIC)~\cite{unterthiner2019fvd} are reported to further assess the perceptual consistency between the input image and the animation results. 
The PIC can be calculated by $\frac{1}{L}  {\textstyle \sum_{i=0}^{L-1}} (1-D(\mathcal{I},x_{i}))$, where $\mathcal{I},x_{i}, L$ are input image, video frames, and video length, respectively. $D(\cdot, \cdot)$ denotes perceptual distance metric  DreamSim~\cite{fu2023dreamsim}. We measure these metrics at the resolution of 256 $\times$ 256 with 16 frames. As shown in Fig.~\ref{fig:ab masking ratio}, the optimal ratio is surprisingly high. The ratio of 70\% obtains the best performance in two metrics. 
An extremely high mask ratio leads to a decrease in the quality of the generated video due to the weak condition of the input image. 
Also, we compare the visual results of training without first-frame masking and with the optimal masking ratio in ~\cref{fig:ab masking ratio}. 
From the results, we can observe that, without the first-frame masking training, the model fails to learn the correct temporal motion and presents incorrect structures.
We then visualize the reconstruction results of the masked input image and generated video frames in Fig.~\ref{fig:Reconstructions}. 
It can be observed that the first frame can be reasonably reconstructed in the generation process and the generated videos maintain good background consistency with input images.
% even if some details differ from the input image (\textit{e.g.} castle in 4th row). 

% \noindent\textbf{Motion strength guidance.}

% \begin{wraptable}{r}{0.7\linewidth}

% \includegraphics[scale=0.5]{
\begin{figure*}[t]
  % \centering
  \begin{tabular}{c@{\hspace{0.1em}}c@{\hspace{0.1em}}c@{\hspace{0.1em}}c@{\hspace{0.1em}}c@{\hspace{0.1em}}c@{\hspace{0.1em}}c@{\hspace{0.1em}}}
  User Click & W/o D+M & W/o D & W/o M & Ours  \\
  % & {\small I2VGen-XL}~\cite{i2vgenxl} & {\small RegionMaker(ours)}\\
    \includegraphics[width=0.2\linewidth]{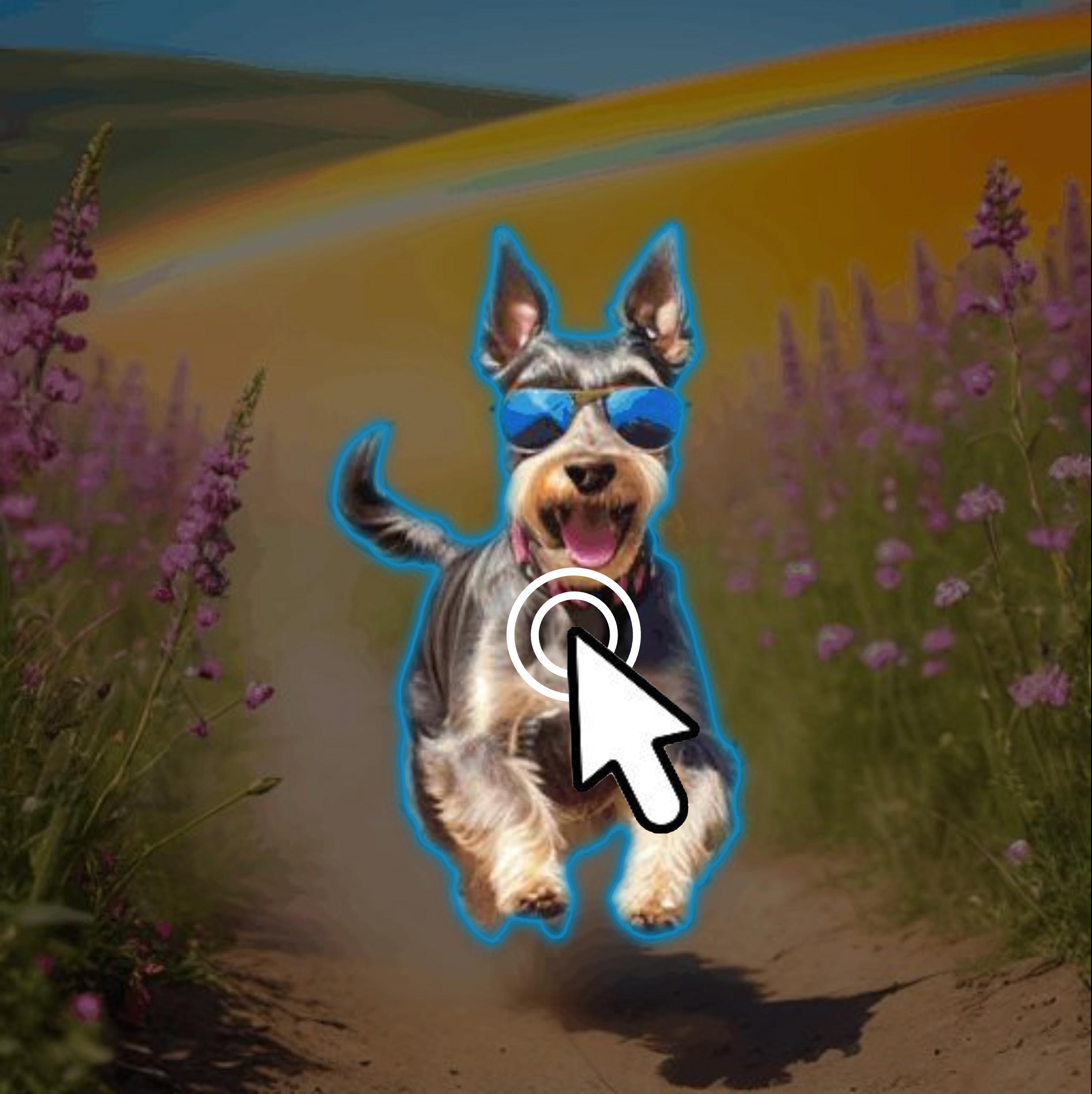} &
    \animategraphics[width=0.2\linewidth]{8}{gif/motion_cross/dog_running/wo_d/frame_}{1}{16}&
        \animategraphics[width=0.2\linewidth]{8}{gif/motion_cross/dog_running/wo_d_m/frame_}{1}{16}&
     \animategraphics[width=0.2\linewidth]{8}{gif/motion_cross/dog_running/wo_m2/frame_}{1}{16}&
         \animategraphics[width=0.2\linewidth]{8}{gif/motion_cross/dog_running/ours/frame_}{1}{16}  \\

    \end{tabular}
    
% \vspace{-0.7em}
  \caption{\textbf{Qualitative results of ablation the constructed short prompt dataset (D) and motion-augmented module (M)}. The motion prompt is \enquote{running}.
  }
  \vspace{-1em}
\label{fig:ab motion} 
\end{figure*}
  % \vspace{-1cm}

\noindent\textbf{Motion-augmented module.}
\begin{figure*}[t]
  % \centering
  \begin{tabular}{c@{\hspace{0.1em}}c@{\hspace{0.1em}}c@{\hspace{0.1em}}c@{\hspace{0.1em}}c@{\hspace{0.1em}}c@{\hspace{0.1em}}c}
       User Click & OFM=4 & OFM=8 & OFM=12 & {OFM=16} \\
  % & {\small I2VGen-XL}~\cite{i2vgenxl} & {\small RegionMaker(ours)}\\
    \includegraphics[width=0.2\linewidth]{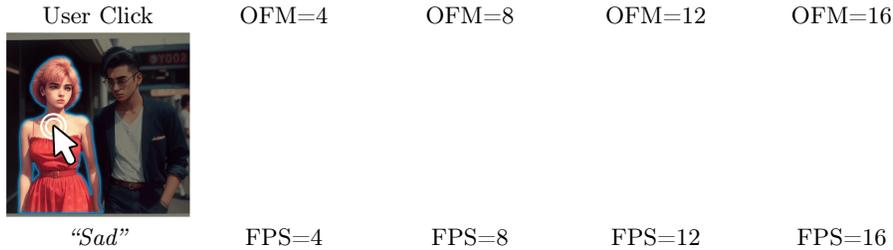} &
    \animategraphics[width=0.2\linewidth]{8}{gif/ablation_motion_cry/OFC_4/frame_}{1}{16}&
     \animategraphics[width=0.2\linewidth]{8}{gif/ablation_motion_cry/OFC_8/frame_}{1}{16}&
     \animategraphics[width=0.2\linewidth]{8}{gif/ablation_motion_cry/FPS_16/frame_}{1}{16}&
     \animategraphics[width=0.2\linewidth]{8}{gif/ablation_motion_cry/OFC_16/frame_}{1}{16}
     \\
      \textit{``Sad''}  & FPS=4 & FPS=8 & FPS=12 & FPS=16 \\
  % & {\small I2VGen-XL}~\cite{i2vgenxl} & {\small RegionMaker(ours)}\\
    &
    \animategraphics[width=0.2\linewidth]{8}{gif/ablation_motion_cry/FPS_4/frame_}{1}{10}&
     \animategraphics[width=0.2\linewidth]{8}{gif/ablation_motion_cry/FPS_8/frame_}{1}{16}&
     \animategraphics[width=0.2\linewidth]{8}{gif/ablation_motion_cry/FPS_12/frame_}{1}{16}&
     \animategraphics[width=0.2\linewidth]{8}{gif/ablation_motion_cry/OFC_8/frame_}{1}{16}\\
    \end{tabular}
    
\vspace{-0.3em}
  \caption{
  \textbf{Comparisons between our optical flow motion magnitude control (OFM) and FPS-based motion magnitude control (FPS)}.
  Our control method can effectively and almost linearly control the motion intensity.
  % Qualitative comparison results between publicly available image-to-video tools and our proposed RegionMaker. 
  {\textit{View with Acrobat Reader to play the animation clips.}}
  % \textbf{Static frames are provided in supplementary materials.}}}
  }
\vspace{-0.3cm}
\label{fig:magnitude} 
\end{figure*}
% }
To investigate the roles of our dataset and motion-augmented (MA) module, we examine two variants: 1) \textbf{Ours w/o D+M}, we apply the basic motion module designed in AnimateDiff~\cite{he2023animate} and finetune the model on WebVid-10M.
2) \textbf{Ours w/o D}, during training stage, we only use public WebVid-10M to optimize the proposed method. The input of MA module is the original prompt from WebVid-10M. 3) \textbf{Ours w/o M}, by removing the MA module. The short motion-related prompts are fed into cross-attention in the spatial module.
We also conduct the qualitative comparison in Fig.~\ref{fig:ab motion}. The performance of \enquote{Ours w/o D+M} declines significantly due to its inability to semantically comprehend the input image without a short prompt, leading to small motion in the generated videos (see the 2nd column). When we remove the MA module, it exhibits limited motion magnitude. 
We report the quantitative ablation study of the designed module in Table.~\ref{tab:ablation motion augmented module} and the same setting as Sec.~\ref{sec:qr} is applied to evaluate the performance comprehensively. 
Eliminating Webvid-Motion finetuning leads to a significant decrease in the FVD and text alignment. 
In contrast, our full method effectively achieves regional image animation with natural motion and coherent frames.

 % leverages the proposed module for

% \subsection{}

\noindent\textbf{Motion magnitude control.}
We present the comparison results in Fig.~\ref{fig:magnitude} for FPS-based and flow-based motion magnitude control, respectively.
We observe that the motion control using FPS is not precise enough. 
For example, the difference between FPS=4 and FPS=8 is not significant (the 2nd row of Fig.~\ref{fig:ab motion}). 
In contrast, optical flow magnitude (OFM) for motion control can effectively manage the intensity of motion. 
From OFM=4 to OFM=16, it is apparent to observe the increase of motion strength about \texttt{\enquote{Sad}}. 
At OFM=16, it's interesting that the girl expresses her sadness by lowering her head and covering her face.

\subsection{Application}
\label{sec:application}
\begin{figure*}[t]
  % \centering
  \begin{tabular}{c@{\hspace{0.1em}}c@{\hspace{0.1em}}c@{\hspace{0.1em}}c@{\hspace{0.1em}}c@{\hspace{0.1em}}c@{\hspace{0.1em}}c}
  User Click & Output &  User Click & Output &  \\
  % & {\small I2VGen-XL}~\cite{i2vgenxl} & {\small RegionMaker(ours)}\\
    \includegraphics[width=0.2\linewidth]{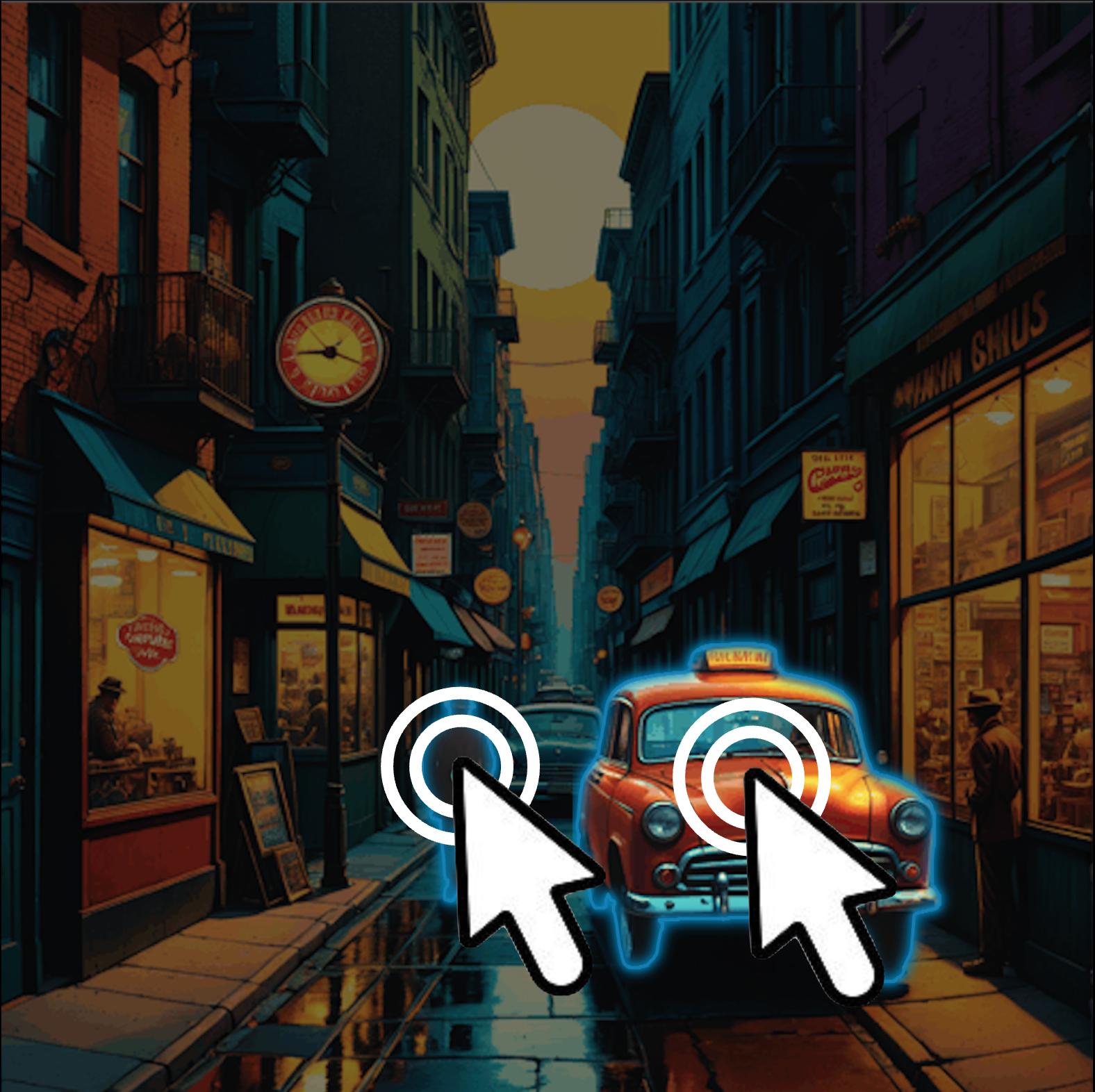} &
    \animategraphics[width=0.2\linewidth]{8}{gif/application/car/frame_}{1}{16}&
     \includegraphics[width=0.2\linewidth]{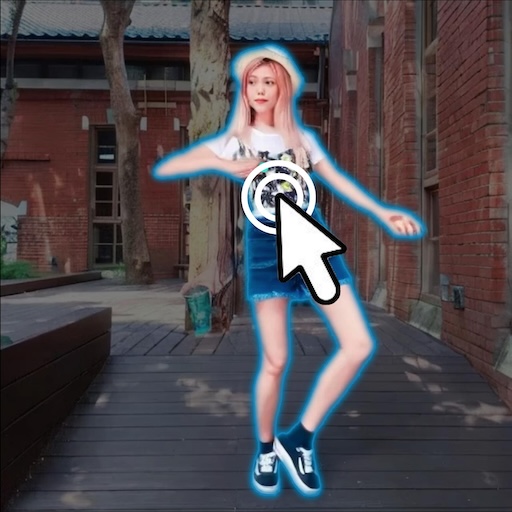}&
     \animategraphics[width=0.4\linewidth]{8}{gif/application/dancing/frame_}{1}{16}
     &
     % % \includegraphics[width=0.14\linewidth]{gif/AZvsOthers/line3/videocrafter/011_0.jpg}&
     % \animategraphics[width=0.2\linewidth]{8}{gif/ablation_motion/OFC_16/frame_}{1}{16}
     \\
     \multicolumn{2}{c}{\textit{``walking, driving''}}  & \multicolumn{3}{c}{\textit{``dancing''}} &  &  \\
    \end{tabular}
    
\vspace{-0.7em}
  \caption{\textbf{The Application of our approach}. Our approach can support multiple regions animation as well as precise motion control such as human pose.
  % \textit{Best viewed with Acrobat Reader. Click the video to play the animation clips. \textbf{Static frames are provided in supplementary materials.}}
  }
  \vspace{-0.3cm}
\label{fig:application} 
\end{figure*}
% }
\noindent\textbf{Multi-regions image animation.}
Using the technology of regional prompter~\cite{regional}, we can achieve multi-region image animation by different short motion prompts. As shown on the left one in Fig.~\ref{fig:application},  we can animate the man and car using \texttt{\enquote{walking, driving}}, respectively. The background of the video is stable, and only selected objects are animated. 

% \noindent\textbf{The control in motion strength.} Thanks to the proposed optical flow-based motion strength control, we can control the motion strength accurately. As shown in the 1st row and 3rd row of Fig.~\ref{fig:ab motion}, we can control the magnitude of the flag and girl's movement by setting various OFM.

\noindent\textbf{Regional image animation with ControlNet~\cite{zhang2023adding}.} In addition, our framework can be combined with ControlNet for conditional regional image animation. 
In the case on the right side of Fig.~\ref{fig:application}, we present the use of pose conditioning for conditional generation. 
It shows that we generate pose-aligned characters with good temporal consistency while maintaining stability of the background.

\section{Limitation}
\begin{figure}[t]
  % \centering
  \begin{minipage}[t]{0.6\textwidth}
      \begin{tabular}{c@{\hspace{0.1em}}c@{\hspace{0.1em}}c@{\hspace{0.1em}}c@{\hspace{0.1em}}c@{\hspace{0.1em}}c@{\hspace{0.1em}}c@{\hspace{0.1em}}}
      User Click &  Output \\
      % & {\small I2VGen-XL}~\cite{i2vgenxl} & {\small RegionMaker(ours)}\\
        \includegraphics[width=0.4\linewidth]{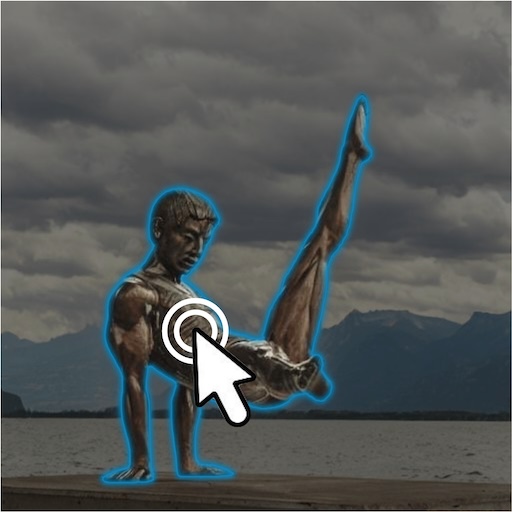} &
        \animategraphics[width=0.4\linewidth]{8}{gif/limitation/frame_}{1}{16}& \\
        \multicolumn{2}{c}{\textit{``Doing a thomas flair''}}
        \end{tabular}
    \end{minipage}
    \begin{minipage}[t]{0.32\textwidth}
        \vspace{-50pt}
        \caption{
        \textbf{Limitation}. 
        Our approach is limited in generating large and complex human motions, as shown in the video. This may be due to the complexity of the action and the rareness of related training samples.
        % \textit{Best viewed with Acrobat Reader. Click the video to play the animation clips. \textbf{Static frames are provided in supplementary materials.}}
        }
        \label{fig:limitation1} 
  \end{minipage}
  \vspace{-0.7cm}
\end{figure}

Although our approach enables click and short motion prompt control,
it still faces the challenge of generating large and complex motion, as shown in Fig.~\ref{fig:limitation1}.
%
% The model fails to generate text-aligned motion while maintaining realistic temporal coherence.
This may be due to the complexity of the motion and the dataset bias, \textit{e.g.}, the training dataset contains limited samples with complex motion.
% These limitations mainly stem from the constraints in motion prior to within the pretrained video generation model. The basic model struggles to generate complex motions, such as large sports movements. As shown in Fig.~\ref{fig:limitation1}, the model fails to generate text-aligned motion while maintaining realistic temporal coherence. However, we believe these limitations can be solved with a better video foundation model with more powerful motion prior.

\section{Conclusion}
\label{sec:conclusion}
In this paper, we present Follow-Your-Click to tackle the problem of generating controllable and local animation. To the best of our knowledge, we are the first I2V framework that is capable of regional image animation via a simple \textit{click} and a \textit{short motion-related prompt}. 
To support this, the promptable segmentation tool SAM is firstly incorporated into our framework for a user-friendly interaction. 
To achieve the short prompt following abilities, we propose a motion-augmented module and a constructed short prompt dataset to achieve this goal. 
To improve the generated temporal motion quality, we propose the first-frame masking strategy which significantly improves the generation performance.
To enable accurate learning of motion speed, we leverage the optical flow score to control the magnitude of motion accurately. 
Our experimental results highlight the effectiveness and superiority of our approach compared to existing baselines.

\section*{Acknowledgments}
We thank Jiaxi Feng, Yabo Zhang, Wenzhe Zhao, Mengyang Liu, Jianbing Wu and Qi Tian for their helpful comments. 
This project was supported by the National Key R\&D Program of China under grant number 2022ZD0161501.

% The paper ends with a conclusion. 

% \clearpage\mbox{}Page \thepage\ of the manuscript.
% \clearpage\mbox{}Page \thepage\ of the manuscript.
% \clearpage\mbox{}Page \thepage\ of the manuscript.
% \clearpage\mbox{}Page \thepage\ of the manuscript.
% \clearpage\mbox{}Page \thepage\ of the manuscript. This is the last page.
% \par\vfill\par
% Now we have reached the maximum length of an ECCV \ECCVyear{} submission (excluding references).
% References should start immediately after the main text, but can continue past p.\ 14 if needed.
% \clearpage  % TODO REVIEW/FINAL: This \clearpage needs to be removed from both review and camera-ready versions.

% ---- Bibliography ----
%
% BibTeX users should specify bibliography style 'splncs04'.
% References will then be sorted and formatted in the correct style.
%
\bibliographystyle{splncs04}
\bibliography{main}

\begin{thebibliography}{10}
\providecommand{\url}[1]{\texttt{#1}}
\providecommand{\urlprefix}{URL }
\providecommand{\doi}[1]{https://doi.org/#1}

\bibitem{gpt4}
Chatgpt-4. \url{https://chat.openai.com/} (2023)

\bibitem{civitai}
Civitai. \url{https://civitai.com/} (2023)

\bibitem{gen2}
Gen-2. \url{https://runwayml.com/ai-magic-tools/gen-2/} (2023)

\bibitem{genmo}
Genmo. \url{https://www.genmo.ai/} (2023)

\bibitem{i2vgenxl}
I2vgen-xl. \url{https://modelscope.cn/models/damo/Image-to-Video/summary} (2023)

\bibitem{pikalabs}
Pika labs. \url{https://www.pika.art/} (2023)

\bibitem{regional}
Regional prompter. \url{https://github.com/hako-mikan/sd-webui-regional-prompter} (2023)

\bibitem{bain2021frozen}
Bain, M., Nagrani, A., Varol, G., Zisserman, A.: Frozen in time: A joint video and image encoder for end-to-end retrieval. In: Proceedings of the IEEE/CVF International Conference on Computer Vision. pp. 1728--1738 (2021)

\bibitem{bertiche2023blowing}
Bertiche, H., Mitra, N.J., Kulkarni, K., Huang, C.H.P., Wang, T.Y., Madadi, M., Escalera, S., Ceylan, D.: Blowing in the wind: Cyclenet for human cinemagraphs from still images. In: Proceedings of the IEEE/CVF Conference on Computer Vision and Pattern Recognition. pp. 459--468 (2023)

\bibitem{blattmann2023stable}
Blattmann, A., Dockhorn, T., Kulal, S., Mendelevitch, D., Kilian, M., Lorenz, D., Levi, Y., English, Z., Voleti, V., Letts, A., et~al.: Stable video diffusion: Scaling latent video diffusion models to large datasets. arXiv preprint arXiv:2311.15127  (2023)

\bibitem{blattmann2021understanding}
Blattmann, A., Milbich, T., Dorkenwald, M., Ommer, B.: Understanding object dynamics for interactive image-to-video synthesis. In: Proceedings of the IEEE/CVF Conference on Computer Vision and Pattern Recognition. pp. 5171--5181 (2021)

\bibitem{blattmann2023align}
Blattmann, A., Rombach, R., Ling, H., Dockhorn, T., Kim, S.W., Fidler, S., Kreis, K.: Align your latents: High-resolution video synthesis with latent diffusion models. In: Proceedings of the IEEE/CVF Conference on Computer Vision and Pattern Recognition. pp. 22563--22575 (2023)

\bibitem{chai2023stablevideo}
Chai, W., Guo, X., Wang, G., Lu, Y.: Stablevideo: Text-driven consistency-aware diffusion video editing. In: Proceedings of the IEEE/CVF International Conference on Computer Vision. pp. 23040--23050 (2023)

\bibitem{chen2023controlavideo}
Chen, W., Wu, J., Xie, P., Wu, H., Li, J., Xia, X., Xiao, X., Lin, L.: Control-a-video: Controllable text-to-video generation with diffusion models (2023)

\bibitem{chen2023livephoto}
Chen, X., Liu, Z., Chen, M., Feng, Y., Liu, Y., Shen, Y., Zhao, H.: Livephoto: Real image animation with text-guided motion control. arXiv preprint arXiv:2312.02928  (2023)

\bibitem{chen2023seine}
Chen, X., Wang, Y., Zhang, L., Zhuang, S., Ma, X., Yu, J., Wang, Y., Lin, D., Qiao, Y., Liu, Z.: Seine: Short-to-long video diffusion model for generative transition and prediction. arXiv preprint arXiv:2310.20700  (2023)

\bibitem{cheng2020time}
Cheng, C.C., Chen, H.Y., Chiu, W.C.: Time flies: Animating a still image with time-lapse video as reference. In: Proceedings of the IEEE/CVF Conference on Computer Vision and Pattern Recognition. pp. 5641--5650 (2020)

\bibitem{cheng2023segment}
Cheng, Y., Li, L., Xu, Y., Li, X., Yang, Z., Wang, W., Yang, Y.: Segment and track anything. arXiv preprint arXiv:2305.06558  (2023)

\bibitem{dai2023animateanything}
Dai, Z., Zhang, Z., Yao, Y., Qiu, B., Zhu, S., Qin, L., Wang, W.: Animateanything: Fine-grained open domain image animation with motion guidance. arXiv e-prints pp. arXiv--2311 (2023)

\bibitem{ding2022cogview2}
Ding, M., Zheng, W., Hong, W., Tang, J.: Cogview2: Faster and better text-to-image generation via hierarchical transformers. Advances in Neural Information Processing Systems  \textbf{35},  16890--16902 (2022)

\bibitem{dorkenwald2021stochastic}
Dorkenwald, M., Milbich, T., Blattmann, A., Rombach, R., Derpanis, K.G., Ommer, B.: Stochastic image-to-video synthesis using cinns. In: Proceedings of the IEEE/CVF Conference on Computer Vision and Pattern Recognition. pp. 3742--3753 (2021)

\bibitem{esser2023structure}
Esser, P., Chiu, J., Atighehchian, P., Granskog, J., Germanidis, A.: Structure and content-guided video synthesis with diffusion models. In: Proceedings of the IEEE/CVF International Conference on Computer Vision. pp. 7346--7356 (2023)

\bibitem{feichtenhofer2022masked}
Feichtenhofer, C., Li, Y., He, K., et~al.: Masked autoencoders as spatiotemporal learners. Advances in neural information processing systems  \textbf{35},  35946--35958 (2022)

\bibitem{fu2023dreamsim}
Fu, S., Tamir, N., Sundaram, S., Chai, L., Zhang, R., Dekel, T., Isola, P.: Dreamsim: Learning new dimensions of human visual similarity using synthetic data. arXiv preprint arXiv:2306.09344  (2023)

\bibitem{gao2023imperceptible}
Gao, K., Bai, J., Wu, B., Ya, M., Xia, S.T.: Imperceptible and robust backdoor attack in 3d point cloud. IEEE Transactions on Information Forensics and Security  \textbf{19},  1267--1282 (2023)

\bibitem{gao2024inducing}
Gao, K., Bai, Y., Gu, J., Xia, S.T., Torr, P., Li, Z., Liu, W.: Inducing high energy-latency of large vision-language models with verbose images. In: ICLR (2024)

\bibitem{geng2018warp}
Geng, J., Shao, T., Zheng, Y., Weng, Y., Zhou, K.: Warp-guided gans for single-photo facial animation. ACM Transactions on Graphics (ToG)  \textbf{37}(6),  1--12 (2018)

\bibitem{guo2023i2v}
Guo, X., Zheng, M., Hou, L., Gao, Y., Deng, Y., Ma, C., Hu, W., Zha, Z., Huang, H., Wan, P., et~al.: I2v-adapter: A general image-to-video adapter for video diffusion models. arXiv preprint arXiv:2312.16693  (2023)

\bibitem{guo2023sparsectrl}
Guo, Y., Yang, C., Rao, A., Agrawala, M., Lin, D., Dai, B.: Sparsectrl: Adding sparse controls to text-to-video diffusion models (2023)

\bibitem{guo2023animatediff}
Guo, Y., Yang, C., Rao, A., Wang, Y., Qiao, Y., Lin, D., Dai, B.: Animatediff: Animate your personalized text-to-image diffusion models without specific tuning. arXiv preprint arXiv:2307.04725  (2023)

\bibitem{he2023camouflaged}
He, C., Li, K., Zhang, Y., Tang, L., Zhang, Y., Guo, Z., Li, X.: Camouflaged object detection with feature decomposition and edge reconstruction. In: Proceedings of the IEEE/CVF Conference on Computer Vision and Pattern Recognition. pp. 22046--22055 (2023)

\bibitem{he2023weaklysupervised}
He, C., Li, K., Zhang, Y., Xu, G., Tang, L., Zhang, Y., Guo, Z., Li, X.: Weakly-supervised concealed object segmentation with sam-based pseudo labeling and multi-scale feature grouping. arXiv preprint arXiv:2305.11003  (2023)

\bibitem{he2024strategic}
He, C., Li, K., Zhang, Y., Zhang, Y., Guo, Z., Li, X., Danelljan, M., Yu, F.: Strategic preys make acute predators: Enhancing camouflaged object detectors by generating camouflaged objects (2024)

\bibitem{he2022masked}
He, K., Chen, X., Xie, S., Li, Y., Doll{\'a}r, P., Girshick, R.: Masked autoencoders are scalable vision learners. In: Proceedings of the IEEE/CVF conference on computer vision and pattern recognition. pp. 16000--16009 (2022)

\bibitem{he2023animate}
He, Y., Xia, M., Chen, H., Cun, X., Gong, Y., Xing, J., Zhang, Y., Wang, X., Weng, C., Shan, Y., et~al.: Animate-a-story: Storytelling with retrieval-augmented video generation. arXiv preprint arXiv:2307.06940  (2023)

\bibitem{he2022latent}
He, Y., Yang, T., Zhang, Y., Shan, Y., Chen, Q.: Latent video diffusion models for high-fidelity video generation with arbitrary lengths. arXiv preprint arXiv:2211.13221  (2022)

\bibitem{hinz2021improved}
Hinz, T., Fisher, M., Wang, O., Wermter, S.: Improved techniques for training single-image gans. In: Proceedings of the IEEE/CVF Winter Conference on Applications of Computer Vision. pp. 1300--1309 (2021)

\bibitem{ho2022imagen}
Ho, J., Chan, W., Saharia, C., Whang, J., Gao, R., Gritsenko, A., Kingma, D.P., Poole, B., Norouzi, M., Fleet, D.J., et~al.: Imagen video: High definition video generation with diffusion models. arXiv preprint arXiv:2210.02303  (2022)

\bibitem{ho2022classifier}
Ho, J., Salimans, T.: Classifier-free diffusion guidance. arXiv preprint arXiv:2207.12598  (2022)

\bibitem{ho2022video}
Ho, J., Salimans, T., Gritsenko, A., Chan, W., Norouzi, M., Fleet, D.J.: Video diffusion models. arXiv:2204.03458  (2022)

\bibitem{holynski2021animating}
Holynski, A., Curless, B.L., Seitz, S.M., Szeliski, R.: Animating pictures with eulerian motion fields. In: Proceedings of the IEEE/CVF Conference on Computer Vision and Pattern Recognition. pp. 5810--5819 (2021)

\bibitem{hong2022cogvideo}
Hong, W., Ding, M., Zheng, W., Liu, X., Tang, J.: Cogvideo: Large-scale pretraining for text-to-video generation via transformers. arXiv preprint arXiv:2205.15868  (2022)

\bibitem{jhou2015animating}
Jhou, W.C., Cheng, W.H.: Animating still landscape photographs through cloud motion creation. IEEE Transactions on Multimedia  \textbf{18}(1),  4--13 (2015)

\bibitem{karras2023dreampose}
Karras, J., Holynski, A., Wang, T.C., Kemelmacher-Shlizerman, I.: Dreampose: Fashion image-to-video synthesis via stable diffusion. arXiv preprint arXiv:2304.06025  (2023)

\bibitem{karras2020analyzing}
Karras, T., Laine, S., Aittala, M., Hellsten, J., Lehtinen, J., Aila, T.: Analyzing and improving the image quality of stylegan. In: Proceedings of the IEEE/CVF conference on computer vision and pattern recognition. pp. 8110--8119 (2020)

\bibitem{li2023generative}
Li, Z., Tucker, R., Snavely, N., Holynski, A.: Generative image dynamics. arXiv preprint arXiv:2309.07906  (2023)

\bibitem{loshchilov2017decoupled}
Loshchilov, I., Hutter, F.: Decoupled weight decay regularization. arXiv preprint arXiv:1711.05101  (2017)

\bibitem{ma2023magicstick}
Ma, Y., Cun, X., He, Y., Qi, C., Wang, X., Shan, Y., Li, X., Chen, Q.: Magicstick: Controllable video editing via control handle transformations. arXiv preprint arXiv:2312.03047  (2023)

\bibitem{ma2023follow}
Ma, Y., He, Y., Cun, X., Wang, X., Shan, Y., Li, X., Chen, Q.: Follow your pose: Pose-guided text-to-video generation using pose-free videos. arXiv preprint arXiv:2304.01186  (2023)

\bibitem{ma2022simvtp}
Ma, Y., Yang, T., Shan, Y., Li, X.: Simvtp: Simple video text pre-training with masked autoencoders. arXiv preprint arXiv:2212.03490  (2022)

\bibitem{mallya2022implicit}
Mallya, A., Wang, T.C., Liu, M.Y.: Implicit warping for animation with image sets. Advances in Neural Information Processing Systems  \textbf{35},  22438--22450 (2022)

\bibitem{nichol2021glide}
Nichol, A., Dhariwal, P., Ramesh, A., Shyam, P., Mishkin, P., McGrew, B., Sutskever, I., Chen, M.: Glide: Towards photorealistic image generation and editing with text-guided diffusion models. arXiv preprint arXiv:2112.10741  (2021)

\bibitem{prashnani2017phase}
Prashnani, E., Noorkami, M., Vaquero, D., Sen, P.: A phase-based approach for animating images using video examples. In: Computer Graphics Forum. vol.~36, pp. 303--311. Wiley Online Library (2017)

\bibitem{ramesh2021zero}
Ramesh, A., Pavlov, M., Goh, G., Gray, S., Voss, C., Radford, A., Chen, M., Sutskever, I.: Zero-shot text-to-image generation. In: International Conference on Machine Learning. pp. 8821--8831. PMLR (2021)

\bibitem{rombach2022high}
Rombach, R., Blattmann, A., Lorenz, D., Esser, P., Ommer, B.: High-resolution image synthesis with latent diffusion models. In: Proceedings of the IEEE/CVF conference on computer vision and pattern recognition. pp. 10684--10695 (2022)

\bibitem{ronneberger2015u}
Ronneberger, O., Fischer, P., Brox, T.: U-net: Convolutional networks for biomedical image segmentation. In: Medical Image Computing and Computer-Assisted Intervention--MICCAI 2015: 18th International Conference, Munich, Germany, October 5-9, 2015, Proceedings, Part III 18. pp. 234--241. Springer (2015)

\bibitem{saharia2022photorealistic}
Saharia, C., Chan, W., Saxena, S., Li, L., Whang, J., Denton, E.L., Ghasemipour, K., Gontijo~Lopes, R., Karagol~Ayan, B., Salimans, T., et~al.: Photorealistic text-to-image diffusion models with deep language understanding. Advances in Neural Information Processing Systems  \textbf{35},  36479--36494 (2022)

\bibitem{salimans2022progressive}
Salimans, T., Ho, J.: Progressive distillation for fast sampling of diffusion models. arXiv preprint arXiv:2202.00512  (2022)

\bibitem{shaham2019singan}
Shaham, T.R., Dekel, T., Michaeli, T.: Singan: Learning a generative model from a single natural image. In: Proceedings of the IEEE/CVF international conference on computer vision. pp. 4570--4580 (2019)

\bibitem{shi2024motion}
Shi, X., Huang, Z., Wang, F.Y., Bian, W., Li, D., Zhang, Y., Zhang, M., Cheung, K.C., See, S., Qin, H., et~al.: Motion-i2v: Consistent and controllable image-to-video generation with explicit motion modeling. arXiv preprint arXiv:2401.15977  (2024)

\bibitem{siarohin2021motion}
Siarohin, A., Woodford, O.J., Ren, J., Chai, M., Tulyakov, S.: Motion representations for articulated animation. In: Proceedings of the IEEE/CVF Conference on Computer Vision and Pattern Recognition. pp. 13653--13662 (2021)

\bibitem{song2020denoising}
Song, J., Meng, C., Ermon, S.: Denoising diffusion implicit models. arXiv preprint arXiv:2010.02502  (2020)

\bibitem{soomro2012ucf101}
Soomro, K., Zamir, A.R., Shah, M.: Ucf101: A dataset of 101 human actions classes from videos in the wild. arXiv preprint arXiv:1212.0402  (2012)

\bibitem{teed2020raft}
Teed, Z., Deng, J.: Raft: Recurrent all-pairs field transforms for optical flow. In: Computer Vision--ECCV 2020: 16th European Conference, Glasgow, UK, August 23--28, 2020, Proceedings, Part II 16. pp. 402--419. Springer (2020)

\bibitem{unterthiner2019fvd}
Unterthiner, T., van Steenkiste, S., Kurach, K., Marinier, R., Michalski, M., Gelly, S.: Fvd: A new metric for video generation  (2019)

\bibitem{vaswani2017attention}
Vaswani, A., Shazeer, N., Parmar, N., Uszkoreit, J., Jones, L., Gomez, A.N., Kaiser, {\L}., Polosukhin, I.: Attention is all you need. Advances in neural information processing systems  \textbf{30} (2017)

\bibitem{voleti2022mcvd}
Voleti, V., Jolicoeur-Martineau, A., Pal, C.: Mcvd-masked conditional video diffusion for prediction, generation, and interpolation. Advances in Neural Information Processing Systems  \textbf{35},  23371--23385 (2022)

\bibitem{wang2023gen}
Wang, F.Y., Chen, W., Song, G., Ye, H.J., Liu, Y., Li, H.: Gen-l-video: Multi-text to long video generation via temporal co-denoising. arXiv preprint arXiv:2305.18264  (2023)

\bibitem{wang2024animatelcm}
Wang, F.Y., Huang, Z., Shi, X., Bian, W., Song, G., Liu, Y., Li, H.: Animatelcm: Accelerating the animation of personalized diffusion models and adapters with decoupled consistency learning. arXiv preprint arXiv:2402.00769  (2024)

\bibitem{wang2023modelscope}
Wang, J., Yuan, H., Chen, D., Zhang, Y., Wang, X., Zhang, S.: Modelscope text-to-video technical report. arXiv preprint arXiv:2308.06571  (2023)

\bibitem{wang2023videocomposer}
Wang, X., Yuan, H., Zhang, S., Chen, D., Wang, J., Zhang, Y., Shen, Y., Zhao, D., Zhou, J.: Videocomposer: Compositional video synthesis with motion controllability. arXiv preprint arXiv:2306.02018  (2023)

\bibitem{wang2024videocomposer}
Wang, X., Yuan, H., Zhang, S., Chen, D., Wang, J., Zhang, Y., Shen, Y., Zhao, D., Zhou, J.: Videocomposer: Compositional video synthesis with motion controllability. Advances in Neural Information Processing Systems  \textbf{36} (2024)

\bibitem{wang2022latent}
Wang, Y., Yang, D., Bremond, F., Dantcheva, A.: Latent image animator: Learning to animate images via latent space navigation. arXiv preprint arXiv:2203.09043  (2022)

\bibitem{weng2019photo}
Weng, C.Y., Curless, B., Kemelmacher-Shlizerman, I.: Photo wake-up: 3d character animation from a single photo. In: Proceedings of the IEEE/CVF conference on computer vision and pattern recognition. pp. 5908--5917 (2019)

\bibitem{xiao2023automatic}
Xiao, W., Liu, W., Wang, Y., Ghanem, B., Li, B.: Automatic animation of hair blowing in still portrait photos. In: Proceedings of the IEEE/CVF International Conference on Computer Vision. pp. 22963--22975 (2023)

\bibitem{xiao2023bridging}
Xiao, Y., Luo, Z., Liu, Y., Ma, Y., Bian, H., Ji, Y., Yang, Y., Li, X.: Bridging the gap: A unified video comprehension framework for moment retrieval and highlight detection. arXiv preprint arXiv:2311.16464  (2023)

\bibitem{xing2023make}
Xing, J., Xia, M., Liu, Y., Zhang, Y., Zhang, Y., He, Y., Liu, H., Chen, H., Cun, X., Wang, X., et~al.: Make-your-video: Customized video generation using textual and structural guidance. arXiv preprint arXiv:2306.00943  (2023)

\bibitem{xing2023dynamicrafter}
Xing, J., Xia, M., Zhang, Y., Chen, H., Wang, X., Wong, T.T., Shan, Y.: Dynamicrafter: Animating open-domain images with video diffusion priors. arXiv preprint arXiv:2310.12190  (2023)

\bibitem{xiong2018learning}
Xiong, W., Luo, W., Ma, L., Liu, W., Luo, J.: Learning to generate time-lapse videos using multi-stage dynamic generative adversarial networks. In: Proceedings of the IEEE Conference on Computer Vision and Pattern Recognition. pp. 2364--2373 (2018)

\bibitem{xu2016msr}
Xu, J., Mei, T., Yao, T., Rui, Y.: Msr-vtt: A large video description dataset for bridging video and language. In: Proceedings of the IEEE conference on computer vision and pattern recognition. pp. 5288--5296 (2016)

\bibitem{xue2022advancing}
Xue, H., Hang, T., Zeng, Y., Sun, Y., Liu, B., Yang, H., Fu, J., Guo, B.: Advancing high-resolution video-language representation with large-scale video transcriptions. In: Proceedings of the IEEE/CVF Conference on Computer Vision and Pattern Recognition. pp. 5036--5045 (2022)

\bibitem{yan2021videogpt}
Yan, W., Zhang, Y., Abbeel, P., Srinivas, A.: Videogpt: Video generation using vq-vae and transformers. arXiv preprint arXiv:2104.10157  (2021)

\bibitem{yu2021vector}
Yu, J., Li, X., Koh, J.Y., Zhang, H., Pang, R., Qin, J., Ku, A., Xu, Y., Baldridge, J., Wu, Y.: Vector-quantized image modeling with improved vqgan. arXiv preprint arXiv:2110.04627  (2021)

\bibitem{yu2022scaling}
Yu, J., Xu, Y., Koh, J.Y., Luong, T., Baid, G., Wang, Z., Vasudevan, V., Ku, A., Yang, Y., Ayan, B.K., et~al.: Scaling autoregressive models for content-rich text-to-image generation. arXiv preprint arXiv:2206.10789  (2022)

\bibitem{zhang2023adding}
Zhang, L., Rao, A., Agrawala, M.: Adding conditional control to text-to-image diffusion models (2023)

\bibitem{zhang2023controlvideo}
Zhang, Y., Wei, Y., Jiang, D., Zhang, X., Zuo, W., Tian, Q.: Controlvideo: Training-free controllable text-to-video generation. arXiv preprint arXiv:2305.13077  (2023)

\bibitem{zhang2023pia}
Zhang, Y., Xing, Z., Zeng, Y., Fang, Y., Chen, K.: Pia: Your personalized image animator via plug-and-play modules in text-to-image models  (2023)

\bibitem{zhou2022magicvideo}
Zhou, D., Wang, W., Yan, H., Lv, W., Zhu, Y., Feng, J.: Magicvideo: Efficient video generation with latent diffusion models. arXiv preprint arXiv:2211.11018  (2022)

\end{thebibliography}

\end{document}